%% file: arxiv.tex
\documentclass[10pt,twocolumn,letterpaper]{article}

\usepackage[pagenumbers]{cvpr} 

\input{preamble}

\definecolor{cvprblue}{rgb}{0.21,0.49,0.74}
\usepackage[pagebackref,breaklinks,colorlinks,allcolors=cvprblue]{hyperref}

\def\confName{CVPR}
\def\confYear{2027}

\title{SeMo3D: Unifying 3D Generation, Editing, and Completion\\
via Selective Modeling}
\author{
Xiaokun Sun, 
Jian Yang,
Zhenyu Zhang$^{*}$ \\
Nanjing University \quad $^{*}$Corresponding Author \\
{\tt\small xiaokun\_sun@smail.nju.edu.cn, csjyang@nju.edu.cn, zhangjesse@foxmail.com} \\
Project Page: \url{https://xiaokunsun.github.io/SeMo3D.github.io/}
}

\begin{document}

\twocolumn[{
\renewcommand\twocolumn[1][]{#1}
\maketitle
\vspace{-13mm}
\begin{center}
    \captionsetup{type=figure}
    \includegraphics[width=0.96\textwidth]{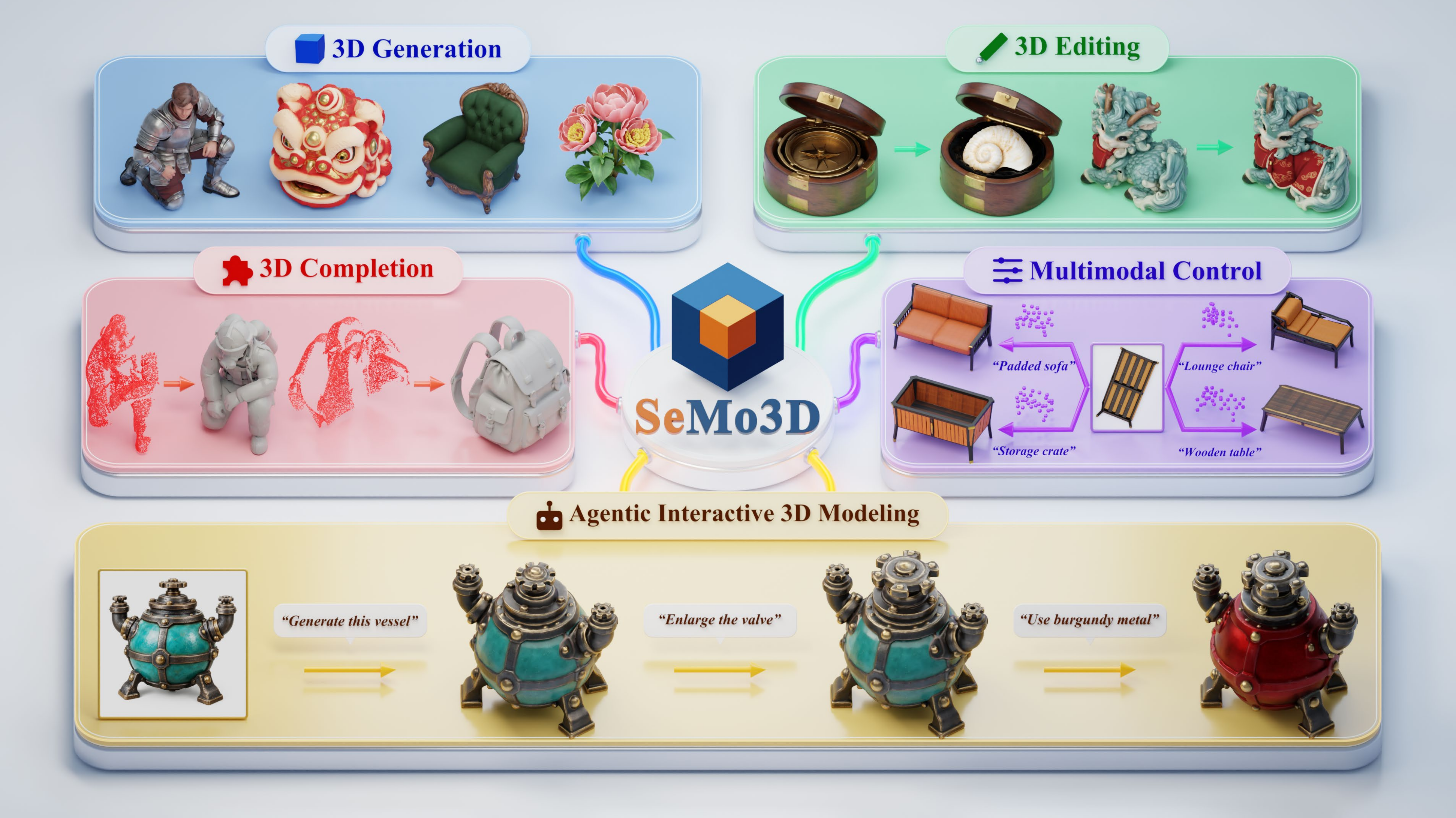}
    \vspace{-3mm}
    \caption{\textbf{SeMo3D} unifies 3D generation, editing, and completion via \textbf{Se}lective \textbf{Mo}deling with image guidance and optional text and points, learning from standard 3D assets without paired editing or completion data. It further supports agentic interactive 3D creation.}
    \label{fig:teaser}
\end{center}
}]

\begin{abstract}
3D generation, editing, and completion are typically treated as regenerating the complete 3D asset under task-specific constraints, even when only a local region needs modification.
This leads to task isolation, inconsistency in content that should remain unchanged, and dependence on task-specific paired data.
We address these issues with a key observation: the three tasks all synthesize a designated region while keeping the rest intact---the full region for generation, the target region for editing, and the missing region for completion.
We term this paradigm Selective Modeling and build SeMo3D upon it; it naturally unifies the three tasks, preserves unchanged content by construction, and enables training without paired editing or completion data, since any 3D asset can be partitioned into a selected region and its context.
For training, we propose region-selective denoising, which randomly selects regions of varying sizes on standard 3D assets and applies noise and loss only to them, jointly learning all three tasks.
For architecture, we propose Omni3DiT, a multi-condition 3D diffusion transformer that distinguishes regions to create from those to retain and fuses image, text, and point guidance for fine-grained control.
Experiments across three benchmarks show SeMo3D surpasses state-of-the-art methods on all tasks, delivering multimodal controllability in generation and strong consistency and accuracy in editing and completion.
SeMo3D further serves as a unified modeling tool for agents, enabling efficient 3D creation through natural-language interaction.
Project page: \url{https://xiaokunsun.github.io/SeMo3D.github.io/}.

\begin{figure*}[tp]
    \centering
    \includegraphics[width=0.90\linewidth]{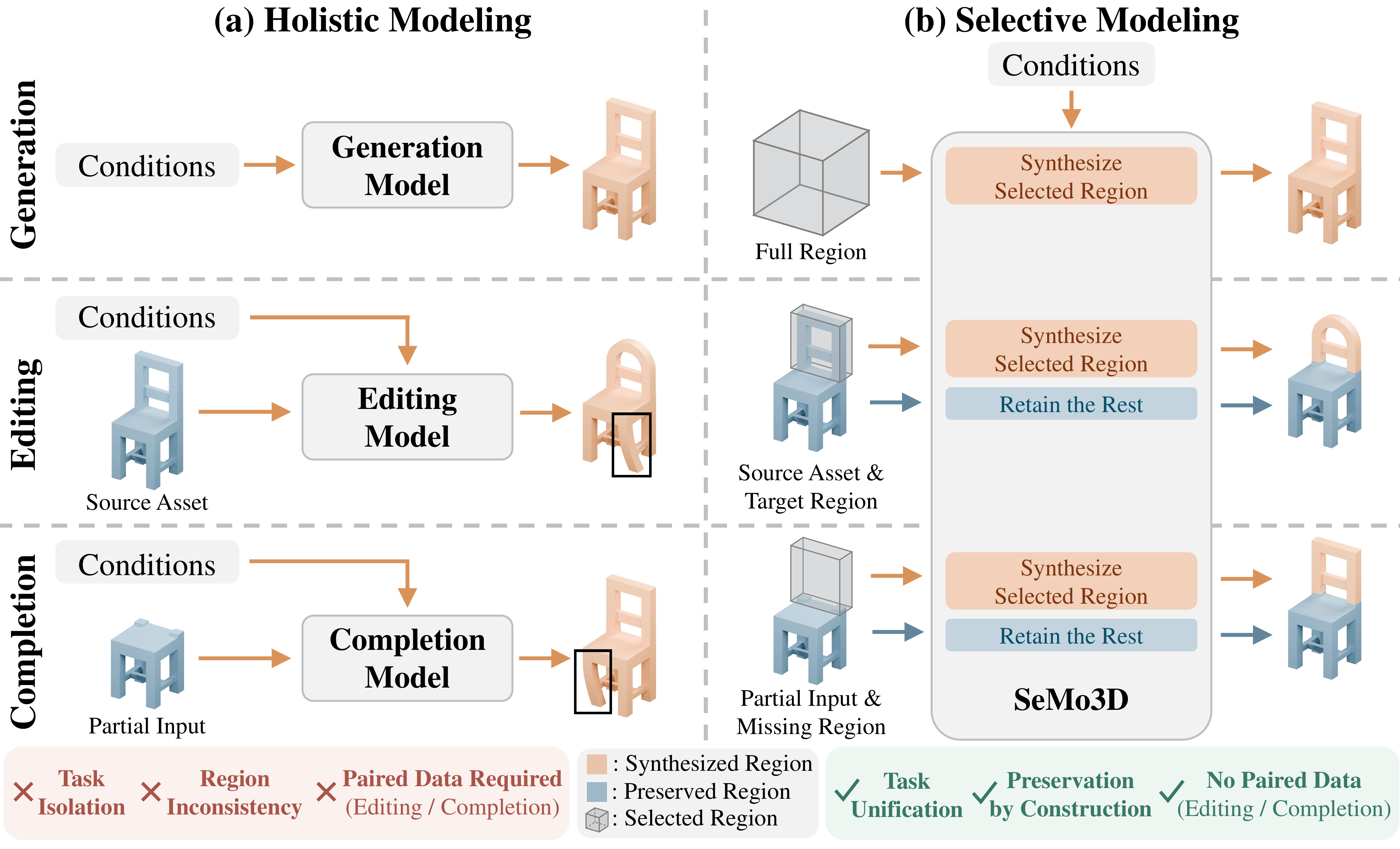} 
    \vspace{-4mm}
    \caption{(a)~Existing \emph{holistic modeling} regenerates the complete asset even for local modification, requiring per-task models and paired editing/completion data, while regions that should remain unchanged are often altered (black boxes). (b)~Our \emph{selective modeling} synthesizes only the designated region, unifying the three tasks, preserving the rest by construction, and needing no task-specific paired data.}
    \label{fig:para}
    \vspace{-5mm}
\end{figure*}

\end{abstract}

\vspace{-5mm}
\section{Introduction}
\label{sec:1}

3D content creation---spanning generation~\cite{3dgen_survey1,3dgen_survey2}, editing~\cite{3dedit_survey}, completion~\cite{3dcomp_survey}, and their combinations---is central to games~\cite{game}, AR/VR~\cite{arvr1,arvr2}, robotics~\cite{robot1,robot2}, and digital twins~\cite{digital}.
Although typically studied in isolation, these tasks are conventionally viewed through the same lens: regenerating a complete 3D asset under task-specific constraints---matching the condition for generation, and additionally the source asset for editing or the partial input for completion.
A unified framework covering all three within one model is thus appealing yet remains an open problem.

Driven by generative models~\cite{diffusion_ddpm,flow} and large-scale 3D datasets~\cite{objaverse,objaverse_xl}, all three tasks have advanced rapidly.
As illustrated in \cref{fig:para}(a), mainstream feed-forward approaches~\cite{trellis,trellis2,3deditformer,partflow,diffcomplete,scdiff} share a common paradigm we term \emph{holistic modeling}: creating the \emph{complete} 3D asset under image or text guidance.
To preserve consistency with the source, editing and completion methods~\cite{3deditformer,partflow,diffcomplete,scdiff} additionally condition on the source asset or partial input, yet they still regenerate the entire object even when only a local region needs to be modified.\footnote{In this paper, editing refers to local operations like addition, deletion, or replacement, not global operations like style transfer.}
This creates a fundamental mismatch between \emph{what needs to change} and \emph{what is actually modeled}, which leads to three key limitations.
\textbf{(1)~Task isolation.} As each task imposes a different constraint, generation~\cite{trellis,3DShape2VecSet,trellis2}, editing~\cite{3deditformer,steer3d,partflow}, and completion~\cite{pdr,diffcomplete,axolotl3d} rely on separate models with distinct architectures and objectives, duplicating design and training effort across tasks and multiplying deployment costs.
\textbf{(2)~Region inconsistency.} Because the full asset is regenerated, source fidelity can only be encouraged rather than guaranteed, and regions that should remain intact are frequently altered (black boxes in \cref{fig:para}(a)).
\textbf{(3)~Editing/completion data dependency.} Regenerating a whole asset that respects the source demands source--target supervision, so many editing~\cite{3deditformer,editverse3d,partflow,shapeup} and completion~\cite{pdr,diffcomplete,axolotl3d} methods require costly large-scale paired datasets (e.g., PartFlow~\cite{partflow} curates 100K editing pairs via a seven-stage pipeline), limiting scalability.
Moreover, most methods accept only a single guidance modality (image or text), offering limited control: text lacks geometric precision, while images leave occluded regions ambiguous.

Our key observation is that, despite their different constraints, the three tasks share a common structure: synthesizing a designated region while keeping the rest intact---the full, target, and missing regions for generation, editing, and completion.
This motivates \emph{selective modeling}: given an asset represented by spatially aligned tokens (e.g., structured latents~\cite{trellis,trellis2}), the model synthesizes only the selected tokens and reuses the rest (\cref{fig:para}(b)), unlike holistic modeling.
Selective modeling thus (1)~unifies the three tasks, which differ only in the selected region; (2)~preserves unchanged content by construction, as context tokens are never regenerated;\footnote{In this paper, preservation ``by construction'' holds at the token level: context tokens stay exactly unchanged. Given the near-lossless reconstruction of structured latents~\cite{trellis2}, this virtually equals asset-level preservation.} and (3)~requires no task-specific paired data, since any ordinary asset can be partitioned into a selected region and its context. This resolves the three limitations above in a single paradigm.

Based on this insight, we present \textbf{SeMo3D}, a multi-condition model unifying 3D generation, editing, and completion via \textbf{Se}lective \textbf{Mo}deling, as illustrated in \cref{fig:teaser}.
We instantiate selective modeling through two key components.
On the training side, we propose \emph{region-selective denoising}: randomly selecting regions of varying sizes on ordinary 3D assets, adding noise only to the selected tokens, and denoising them while keeping the rest clean.
As the selected region ranges from a small portion to the full region, this single procedure spans the entire task spectrum, from local editing through partial completion to full generation, without task-specific paired supervision.
On the architecture side, we design \emph{Omni3DiT}, a unified multi-condition 3D diffusion transformer that jointly processes clean context and noisy selected tokens, and further fuses complementary image, text, and point conditions for fine-grained control.
Experiments on three benchmarks show that SeMo3D outperforms specialized state-of-the-art methods on all tasks, with flexible multi-condition control in generation and high-fidelity preservation in editing and completion.
SeMo3D further supports agentic and interactive 3D modeling through natural-language interaction.

Our contributions are summarized as follows:
\begin{itemize}
    \item We formulate selective modeling, a unified paradigm that casts 3D generation, editing, and completion as synthesizing the designated region while preserving the rest, and present SeMo3D to instantiate this paradigm.
    \item We introduce region-selective denoising, which learns all three tasks from standard 3D datasets alone, without costly large-scale paired editing or completion datasets.
    \item We develop Omni3DiT, a multi-condition 3D diffusion transformer integrating clean 3D context with image, text, and point guidance for unified controllable 3D creation.
\end{itemize}

\section{Related Work}
\label{sec:2}

\textbf{3D Generation.}
Driven by 2D generative models~\cite{diffusion_ddpm,diffusion_sde,diffusion_ddim,flow,ldm,dit}, 3D generation has undergone rapid paradigm shifts.
Early methods~\cite{dreamfusion,mvdream,prolificdreamer,fantasia3d,richdreamer} use score distillation sampling (SDS) to lift 2D priors to 3D, but suffer from long optimization times.
Multi-view 2D methods~\cite{syncdreamer,wonder3d,unique3d,lgm,instantmesh} accelerate generation by first synthesizing multi-view images and then producing 3D shapes in a feed-forward manner.
More recently, native 3D generative models~\cite{3DShape2VecSet,clay,trellis,trellis2,hunyuan3d_2,hunyuan3d_21,hi3dgen,direct3d_s2,triposf,pixal3d} operating directly on 3D latents have pushed quality and scalability further.
Despite this progress, most methods condition on a single modality (image or text), limiting fine-grained controllability.
Recent efforts incorporate pairwise combinations such as image-text~\cite{tigon,know3d} or image-point~\cite{points3d,hunyuan3d_omni}, but none of the above jointly leverages all three modalities to combine the visual detail of images, the semantic expressiveness of text, and the geometric precision of points.

\noindent
\textbf{3D Editing and Completion.}
3D editing and completion have evolved analogously: from SDS-based optimization~\cite{vox,dreameditor,focaldreamer,gaussianeditor,tip,pccomp,compc} through multi-view 2D methods~\cite{instant3dit,cmd,editp23,masklrm,preditor3d,pcdreamer} to native 3D approaches~\cite{3deditformer,partflow,axolotl3d,pdr,scdiff,diffcomplete}.
Native 3D methods are training-free or training-based.
Training-free methods~\cite{voxhammer,nano3d,anchorflow,easy3e,genpc,lascomp} repurpose pretrained generators via inversion or optimization; they largely maintain source consistency but require slow, multi-step pipelines.
Training-based methods~\cite{3deditformer,steer3d,partflow,full,bve,shapeup,editverse3d,prioredit3d,pdr,diffcomplete,scdiff,sgcdiff,axolotl3d} enable fast feed-forward inference but rely on large paired editing or completion datasets (e.g., 100K pairs in PartFlow~\cite{partflow}, 116K in 3DEditFormer~\cite{3deditformer}) and often alter regions that should remain unchanged.
Moreover, both lines mostly use a single modality, limiting controllability.

\noindent
\textbf{Unified 3D Multimodal Models.}
Recent works~\cite{shapellm_omni,universe3d,eva01,omni123,hunyuan3d_bf} consolidate multimodal output capabilities into one model, typically unifying 3D understanding with generation and editing~\cite{shapellm_omni,universe3d,eva01,hunyuan3d_bf} or bridging 2D and 3D generation~\cite{omni123}.
However, these models are largely built on VLM-based architectures, limiting their native 3D modeling capacity, and their notion of ``unity'' centers on merging \emph{different output modalities} (e.g., text, image, and 3D) rather than \emph{different 3D creation sub-tasks} (generation, editing, and completion).
They also typically condition on only one or two modalities, and those that support editing still rely on large-scale paired editing supervision.

Compared with all three lines of work above, SeMo3D offers a distinct solution: it recasts 3D generation, editing, and completion as a single selective-modeling problem formulated directly on native 3D representations, unifying the three tasks without task-specific architectures or per-task paired supervision while preserving the remaining content by construction.
Additionally, SeMo3D supports joint image, text, and point conditioning, enhancing controllability.

\begin{figure*}[tp]
    \centering
    \includegraphics[width=0.90\linewidth]{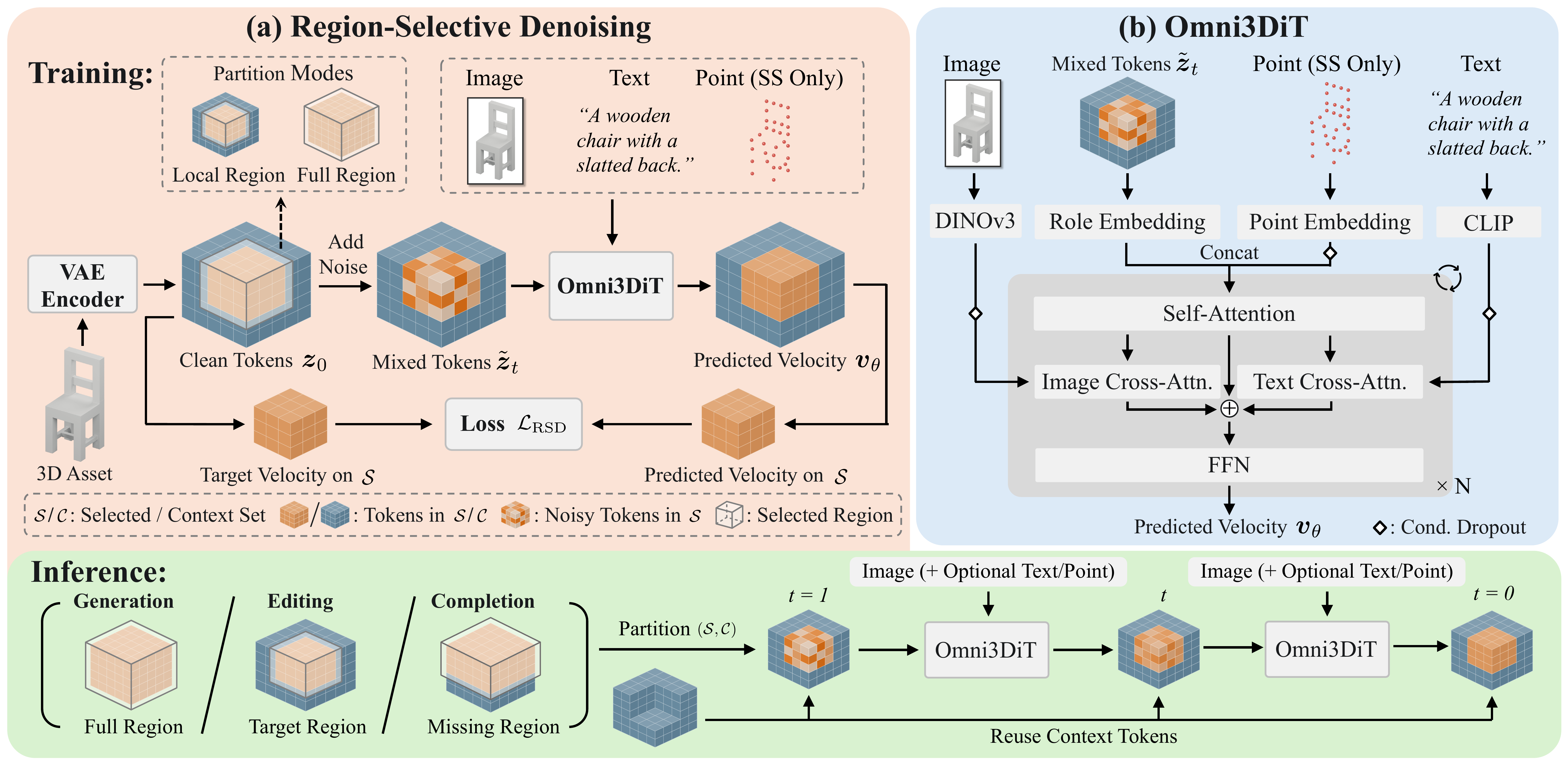}
    \vspace{-4mm}
    \caption{Overview of SeMo3D. (a)~Region-selective denoising. Training applies noise and loss only to a randomly selected set $\mathcal{S}$ while keeping the context set $\mathcal{C}$ clean; as $\mathcal{S}$ ranges from a local box to the full region, one procedure covers editing, completion, and generation. Inference denoises $\mathcal{S}$ from pure noise with a task-defined partition while reusing clean $\mathcal{C}$ at every step. (b)~Omni3DiT jointly processes clean context and noisy selected tokens, fusing image guidance with optional text and point conditions for fine-grained controllable synthesis.}
    \label{fig:pipeline}
    \vspace{-5mm}
\end{figure*}

\section{Methodology}
\label{sec:3}

\subsection{Preliminaries}
\label{sec:3.1}

\textbf{Structured Latents.}
Following Trellis2~\cite{trellis2}, we represent 3D assets as spatially structured latent tokens through a three-stage encoding.
The first stage, sparse structure (SS), encodes a 3D asset into a \emph{dense} latent grid $\mathbf{Z}^{\text{ss}} \in \mathbb{R}^{N^3 \times d_s}$ covering all $N^3$ voxels, from which an SS decoder recovers the occupied voxel set $\mathcal{O}$.
The second and third stages, collectively termed structured latent (SLat) stages, produce \emph{sparse} tokens only at positions in $\mathcal{O}$: shape tokens $\mathbf{Z}^{\text{shape}} \in \mathbb{R}^{|\mathcal{O}| \times d_g}$ encoding fine geometry, and texture tokens $\mathbf{Z}^{\text{tex}} \in \mathbb{R}^{|\mathcal{O}| \times d_t}$ encoding PBR materials conditioned on $\mathbf{Z}^{\text{shape}}$.
Each token at every stage corresponds to a localized spatial region, making the representation amenable to spatially selective operations.

\noindent\textbf{Rectified Flow Matching.}
Each stage employs a rectified flow model~\cite{flow,flowgen} to generate its respective latent tokens.
Given data $\boldsymbol{x}_0$ and noise $\boldsymbol{\epsilon}\!\sim\!\mathcal{N}(\mathbf{0},\mathbf{I})$, the forward process linearly interpolates between them as $\boldsymbol{x}_t = (1{-}t)\,\boldsymbol{x}_0 + t\,\boldsymbol{\epsilon}$ for $t\!\in\![0,1]$.
A velocity network $\boldsymbol{v}_\theta$ is trained by minimizing the conditional flow matching (CFM) objective:
\begin{equation}
    \mathcal{L}_{\text{CFM}} = \mathbb{E}_{t,\,\boldsymbol{x}_0,\,\boldsymbol{\epsilon}}\lVert \boldsymbol{v}_\theta(\boldsymbol{x}_t, t, \boldsymbol{c}) - (\boldsymbol{\epsilon} - \boldsymbol{x}_0)\rVert_2^2,
    \label{eq:flow}
\end{equation}
where $\boldsymbol{c}$ denotes the input conditions, and samples are drawn by integrating the learned ODE from $t{=}1$ (pure noise) to $t{=}0$ (data).
All three stages share this formulation; they differ only in $\boldsymbol{x}_0$ ($\mathbf{Z}^{\text{ss}}$, $\mathbf{Z}^{\text{shape}}$, or $\mathbf{Z}^{\text{tex}}$, respectively) and in the model that operates on each.

\subsection{Overview}
\label{sec:3.2}

\cref{fig:pipeline} illustrates the overall framework.
For a given stage, let $\mathbf{Z}=\{\boldsymbol{z}_i\}_{i\in\mathcal{I}}$ denote its structured latent tokens, where $\mathcal{I}$ indexes all token positions (all $N^3$ voxels for SS; the occupied set $\mathcal{O}$ for SLat).
We partition $\mathcal{I}$ into a \emph{selected set} $\mathcal{S}$ (to be synthesized) and a \emph{context set} $\mathcal{C}$ (to be preserved), where $\mathcal{S} \cup \mathcal{C} = \mathcal{I}$ and $\mathcal{S} \cap \mathcal{C} = \emptyset$.
The model denoises only the tokens in $\mathcal{S}$ while keeping tokens in $\mathcal{C}$ clean.
The three tasks differ solely in how this partition is defined: generation selects the full region ($\mathcal{S}{=}\mathcal{I}$, $\mathcal{C}{=}\emptyset$); editing, the target region; and completion, the missing region.

SeMo3D instantiates this paradigm at each of the three stages, with a dedicated flow model per stage.
For training, \emph{region-selective denoising} (\cref{sec:3.3}, \cref{fig:pipeline}(a)) randomly partitions tokens of ordinary 3D assets and applies noise and loss only to $\mathcal{S}$, so one procedure spans all three tasks without per-task paired supervision.
For the architecture, \emph{Omni3DiT} (\cref{sec:3.4}, \cref{fig:pipeline}(b)) jointly processes clean context tokens and noisy selected tokens, fusing image, text, and point guidance for controllable synthesis.
At inference, image guidance is required; text and sparse points are independently optional.

\begin{figure*}[tp]
    \centering
    \includegraphics[width=0.98\linewidth]{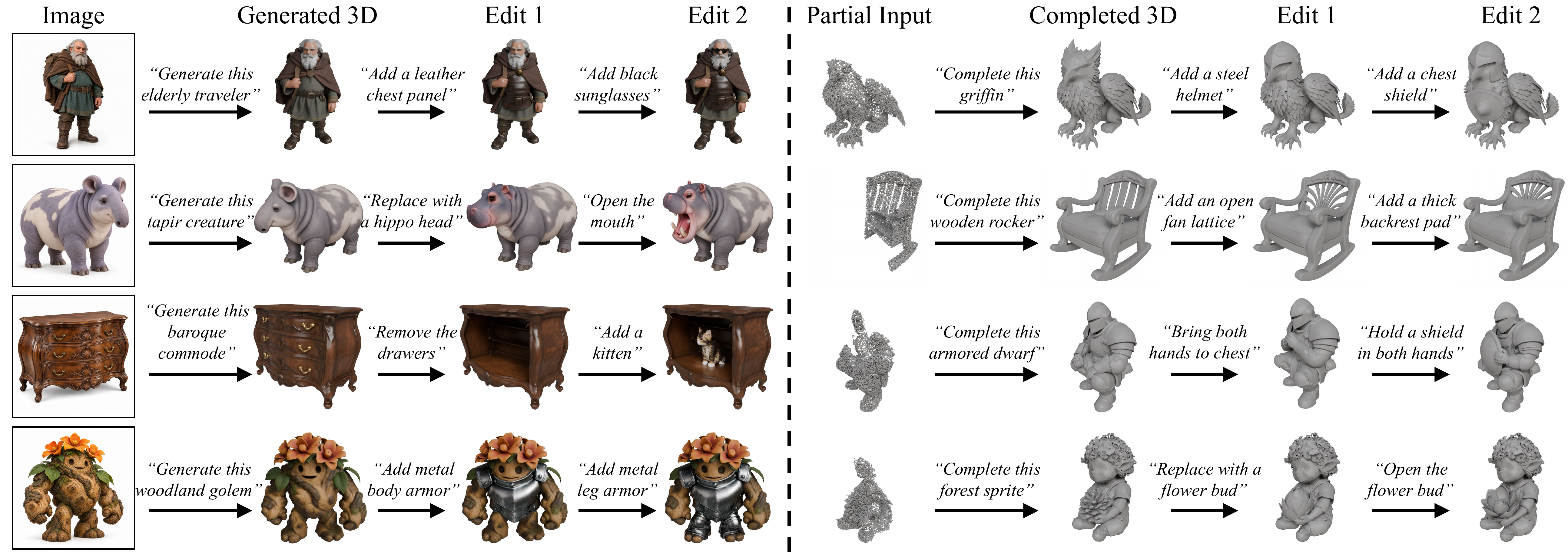}
    \vspace{-3.5mm}
    \caption{Agentic 3D modeling with SeMo3D. Agent prepares input images, determines edit regions, and invokes the appropriate SeMo3D mode in response to requests. Left: generation then editing. Right: completion then editing. \textbf{Refer to Supp.\ Mat. for video results.}}
    \label{fig:agent_show}
    \vspace{-5mm}
\end{figure*}

\subsection{Region-Selective Denoising}
\label{sec:3.3}

As shown in \cref{fig:pipeline}(a), unlike the standard objective (\cref{eq:flow}) that treats all tokens uniformly, region-selective denoising applies noise and loss only to the selected set~$\mathcal{S}$ while keeping context tokens in~$\mathcal{C}$ intact.
We describe the training and inference procedures below.

\noindent\textbf{Training.}
Given the tokens $\mathbf{Z}$ of an ordinary 3D asset and a partition $(\mathcal{S}, \mathcal{C})$, we construct a mixed input $\tilde{\boldsymbol{z}}_t$ by adding noise only to tokens in $\mathcal{S}$ while keeping tokens in $\mathcal{C}$ clean:
\begin{equation}
    \tilde{\boldsymbol{z}}_{t,i} = \begin{cases} (1{-}t)\,\boldsymbol{z}_{0,i} + t\,\boldsymbol{\epsilon}_i, & i \in \mathcal{S}, \\ \boldsymbol{z}_{0,i}, & i \in \mathcal{C}, \end{cases}
    \label{eq:mixed}
\end{equation}
where $\boldsymbol{z}_{0,i}$ is the clean token at position $i$.
The model takes $\tilde{\boldsymbol{z}}_t$ together with a binary mask distinguishing $\mathcal{S}$ from $\mathcal{C}$, and predicts the velocity field over all positions.
The loss is computed only over $\mathcal{S}$; predictions on $\mathcal{C}$ are discarded:
\begin{equation}
    \mathcal{L}_{\text{RSD}} = \mathbb{E}_{t,\,\boldsymbol{z}_0,\,\boldsymbol{\epsilon},\,\mathcal{S}}\!\frac{1}{|\mathcal{S}|}\sum_{i\in\mathcal{S}}\lVert \boldsymbol{v}_\theta(\tilde{\boldsymbol{z}}_t, t, \boldsymbol{c})_i - (\boldsymbol{\epsilon}_i - \boldsymbol{z}_{0,i})\rVert_2^2.
    \label{eq:rsd}
\end{equation}
Here, $1/|\mathcal{S}|$ normalizes the loss across selected-set sizes.
Each step samples one of two partition modes with equal probability:
(1)~\emph{full-region mode} (generation) selects all positions ($\mathcal{S}{=}\mathcal{I}$, $\mathcal{C}{=}\emptyset$), reducing \cref{eq:rsd} to the CFM objective (\cref{eq:flow}) scaled by $1/|\mathcal{I}|$;
(2)~\emph{local-region mode} (editing/completion) selects tokens within a random 3D box.

\noindent\textbf{Inference.}
Tokens in $\mathcal{S}$ start from pure noise and are denoised via ODE integration; tokens in $\mathcal{C}$ stay clean.
The partition is task-dependent:
for generation, the full region is selected ($\mathcal{S}{=}\mathcal{I}$);
for editing, the user provides a mask of the target region;
for completion, the missing region is either user-specified or derived as the complement of the partial input's occupancy; as the SS stage decides occupancy within $\mathcal{S}$, only the actually missing geometry is synthesized.
The three stages run sequentially: the SS stage denoises $\mathcal{S}$ to produce $\mathcal{O}$; the SLat stages inherit the same partition, restricted to $\mathcal{O}$, for shape and texture synthesis.

Benefiting from training regions ranging from a small portion to the full region, the model covers local editing, partial completion, and full generation in one framework, and, though trained only with boxes, handles free-form target and missing regions (\cref{sec:4.2}).
Tokens in $\mathcal{C}$ are never regenerated, guaranteeing source consistency by construction.
Moreover, all three tasks are learned from standard 3D assets without costly per-task paired supervision.

\begin{figure*}[tp]
    \centering
    \includegraphics[width=0.90\linewidth]{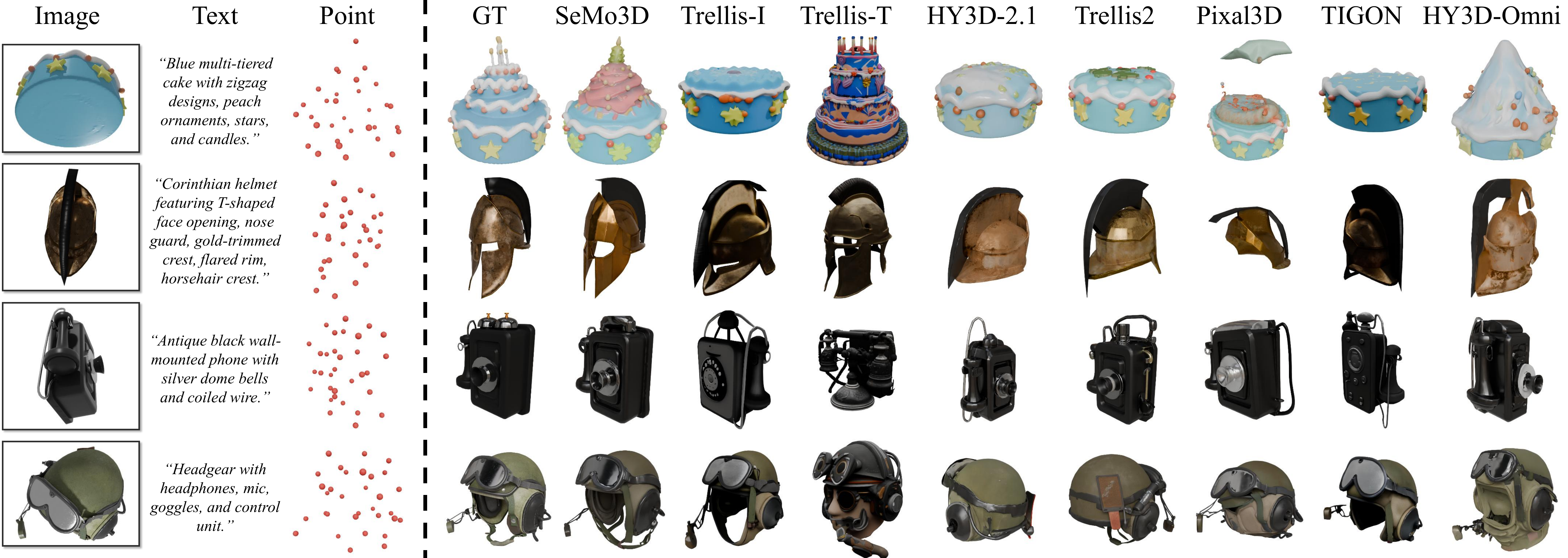}
    \vspace{-4mm}
    \caption{Qualitative comparison of 3D generation on Toys4K~\cite{toys4k}. Only top-performing baselines are shown for brevity. HY3D-Omni~\cite{hunyuan3d_omni} textures are from HY3D-2.1~\cite{hunyuan3d_21}. With three complementary modalities, SeMo3D achieves more controllable and faithful generation.}
    \label{fig:gen}
    \vspace{-5mm}
\end{figure*}

\subsection{Omni3DiT}
\label{sec:3.4}

Omni3DiT extends Trellis2's three-stage image-conditioned 3D DiT~\cite{dit} with four key designs to enable selective modeling with optional text and point guidance (\cref{fig:pipeline}(b)): role embedding, text conditioning, point conditioning, and compositional condition dropout.

\noindent\textbf{Role Embedding.}
In selective modeling, $\tilde{\boldsymbol{z}}_t$ (\cref{eq:mixed}) mixes tokens at different noise levels: tokens in $\mathcal{S}$ carry noise at timestep~$t$, while tokens in $\mathcal{C}$ remain clean.
To help the model distinguish noisy tokens from clean ones, we encode the binary mask of \cref{sec:3.3} as a learned role embedding added after the input projection: tokens in $\mathcal{S}$ receive $\boldsymbol{e}_{\mathcal{S}} \in \mathbb{R}^{d}$ and tokens in $\mathcal{C}$ receive $\boldsymbol{e}_{\mathcal{C}} \in \mathbb{R}^{d}$, where $d$ is the transformer hidden width.
Both embeddings are zero-initialized to preserve pretrained behavior at initialization.

\noindent\textbf{Text Conditioning.}
Trellis2 conditions only on images via DINOv3~\cite{dinov3} cross-attention, yet images cannot describe occluded content.
We add a parallel cross-attention text branch~\cite{know3d,tigon} injecting CLIP~\cite{clip} features.
Both branches share one query and sum their outputs into a residual stream.
The text branch output projection is zero-initialized, leaving the pretrained pathway initially undisturbed.

\noindent\textbf{Point Conditioning.}
Image and text provide rich visual and semantic guidance but cannot specify exact 3D locations.
We therefore add points as a third condition, using sparse rather than dense points since they make user interaction far easier.
In the SS stage, each point is represented by a shared learnable embedding, positioned via RoPE~\cite{rope} at its voxel coordinate, and concatenated into the token sequence~\cite{segvigen}.
Since points share the 3D space of latent tokens, self-attention enables direct spatial reasoning between points and latent tokens, proving substantially more effective than point cross-attention (\cref{sec:4.3}).
The SLat stages omit point conditioning, as image and text suffice for geometry and texture once the sparse structure is determined.

\noindent\textbf{Compositional Condition Dropout.}
Naively finetuning with all conditions active risks letting the pretrained image branch dominate, suppressing the newly added text and point pathways.
At each training step, a random subset of modalities is kept active while the rest are replaced with null embeddings.
The probabilities are set so that the image condition is frequently withheld, compelling the text and point branches to independently carry the full guidance load.

All new modules are zero-initialized or additive, largely preserving pretrained behavior at initialization for stable finetuning.
Together with region-selective denoising, these components enable Omni3DiT to unify generation, editing, and completion under flexible image, text, and point guidance, as validated by ablation studies in \cref{sec:4.3}.

\vspace{-2mm}
\section{Experiments}
\label{sec:4}

Built on these components, SeMo3D supports agentic 3D creation through successive natural-language instructions when integrated with an agent (\cref{fig:agent_show}).
We report key settings and results here; further details are in the Supp.\ Mat.

\subsection{Experimental Settings}
\label{sec:4.1}

\noindent\textbf{Implementation Details.}
We reuse Trellis2's pretrained SC-VAE~\cite{trellis2} and finetune its three-stage flow models at 512 resolution on Trellis-500K~\cite{trellis} with learning rate $5{\times}10^{-5}$ and batch size $32$.
The SS and SLat stages are finetuned for 200K and 100K steps, respectively.
Each training sample pairs one asset with a randomly chosen rendered view and dataset caption as its image and text conditions, with points sampled from the asset's geometry.
Generating a 512-resolution asset takes ${\sim}$5\,s, on par with Trellis2.
For agentic modeling, GPT-6 Astra~\cite{chatgpt} with Codex uses GPT-Image to prepare reference images, determines edit regions, and invokes the appropriate SeMo3D mode.

\noindent\textbf{Baselines.}
We compare with leading methods per task:
(1)~3D generation: Trellis~\cite{trellis}, ShapeLLM-Omni~\cite{shapellm_omni}, HY3D-2.1~\cite{hunyuan3d_21}, Trellis2~\cite{trellis2}, Pixal3D~\cite{pixal3d}, TIGON~\cite{tigon}, and HY3D-Omni~\cite{hunyuan3d_omni}, spanning image, text, image-text, and image-point inputs;
(2)~3D editing: training-free VoxHammer~\cite{voxhammer} and Nano3D~\cite{nano3d}, and training-based 3DEditFormer~\cite{3deditformer} and PartFlow~\cite{partflow};
(3)~3D completion: LaS-Comp~\cite{lascomp} and point-guided HY3D-Omni~\cite{hunyuan3d_omni} fed partial inputs as points.
Baselines receive all supported conditions.

\begin{table}[tp]
    \caption{Quantitative results on 3D generation. Best and second-best in \textbf{bold} and \underline{underlined}. Cond.: I\,=\,Image, T\,=\,Text, P\,=\,Point.}
    \vspace{-3mm}
    \label{tab:gen}
    \centering
    \renewcommand{\arraystretch}{0.8}
    \resizebox{0.95\linewidth}{!}{
        \begin{tabular}{lcccccc}
            \toprule
            Method & Cond. & LPIPS $\downarrow$ & FID $\downarrow$ & KID ($\times$100) $\downarrow$ & DINO $\uparrow$ & CD $\downarrow$ \\
            \midrule
            Trellis~\cite{trellis}       & I    & 0.133 & 6.65 & 0.121 & 0.888 & 0.0413 \\
            Trellis~\cite{trellis}       & T    & 0.183 & 15.48 & 0.202 & 0.795 & 0.0736 \\
            ShapeLLM-Omni~\cite{shapellm_omni}       & I    & 0.167 & 27.15 & 1.167 & 0.751 & 0.0820 \\
            ShapeLLM-Omni~\cite{shapellm_omni}       & T    & 0.185 & 27.30 & 0.836 & 0.760 & 0.0666 \\
            HY3D-2.1~\cite{hunyuan3d_21} & I    & 0.143 & 5.68 & 0.062 & 0.869 & 0.0638 \\
            Trellis2~\cite{trellis2}     & I    & \underline{0.111} & \underline{5.05} & \textbf{0.057} & \underline{0.908} & 0.0330 \\
            Pixal3D~\cite{pixal3d}       & I    & 0.143 & 7.56 & 0.174 & 0.876 & 0.0453 \\
            TIGON~\cite{tigon}           & I, T & 0.127 & 5.90 & 0.092 & 0.897 & 0.0429 \\
            HY3D-Omni~\cite{hunyuan3d_omni} & I, P & 0.121 & 7.18 & 0.107 & 0.879 & \underline{0.0317} \\
            \midrule
            SeMo3D (Ours)                & I, T, P & \textbf{0.079} & \textbf{4.99} & \underline{0.059} & \textbf{0.926} & \textbf{0.0118} \\
            \bottomrule
        \end{tabular}
    }
    \vspace{-5mm}
\end{table}

\noindent\textbf{Benchmarks.}
We evaluate on the public benchmarks Toys4K~\cite{toys4k}, Edit3D-Bench~\cite{voxhammer}, and Omni-Comp~\cite{lascomp} for the three tasks, respectively.
For generation, we report LPIPS~\cite{lpips}, FID~\cite{fid}, KID~\cite{kid}, DINO similarity~\cite{dinov2}, and Chamfer Distance (CD).
For editing, we assess region consistency with CD, PSNR, SSIM~\cite{ssim}, and LPIPS; perceptual quality with FID and FVD; and accuracy with DINO and CLIP similarity.
For completion, we report CD, Earth Mover's Distance (EMD), and F-Score (at a 0.01 threshold).

\noindent\textbf{Evaluation Details.}
For variable-resolution methods, outputs are standardized to 512.
For editing, mask-based methods receive the same ground-truth masks for fair comparison.
For completion, SeMo3D derives its selected region by inverting the partial input's occupancy mask (\cref{sec:3.3}).
Conditions are sampled from the ground truth as in training.

\begin{figure*}[tp]
    \centering
    \includegraphics[width=0.90\linewidth]{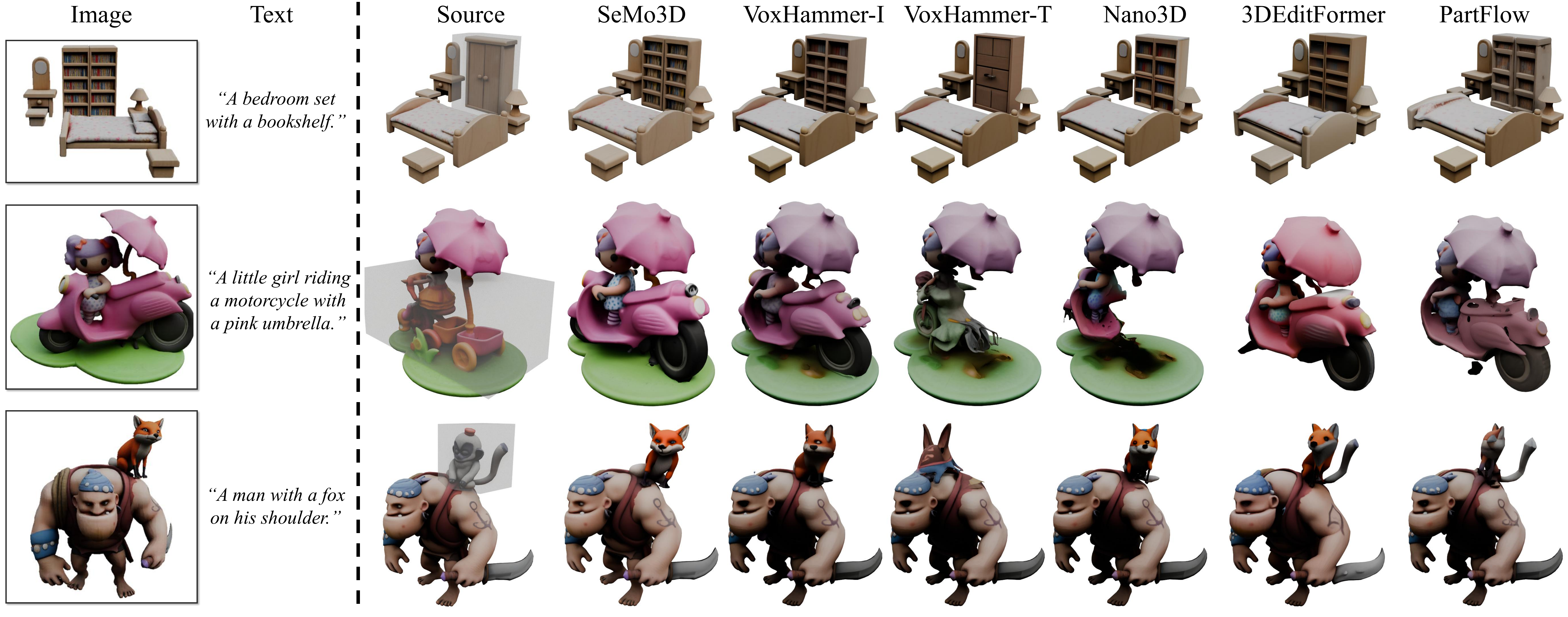}
    \vspace{-4mm}
    \caption{Qualitative comparison of 3D editing on Edit3D-Bench~\cite{voxhammer}. Translucent volumes mark edit regions; points are omitted as Edit3D-Bench lacks them. Without paired editing data, SeMo3D precisely modifies the target region while preserving the rest.}
    \label{fig:edit}
    \vspace{-4mm}
\end{figure*}

\begin{figure*}[tp]
    \centering
    \includegraphics[width=0.90\linewidth]{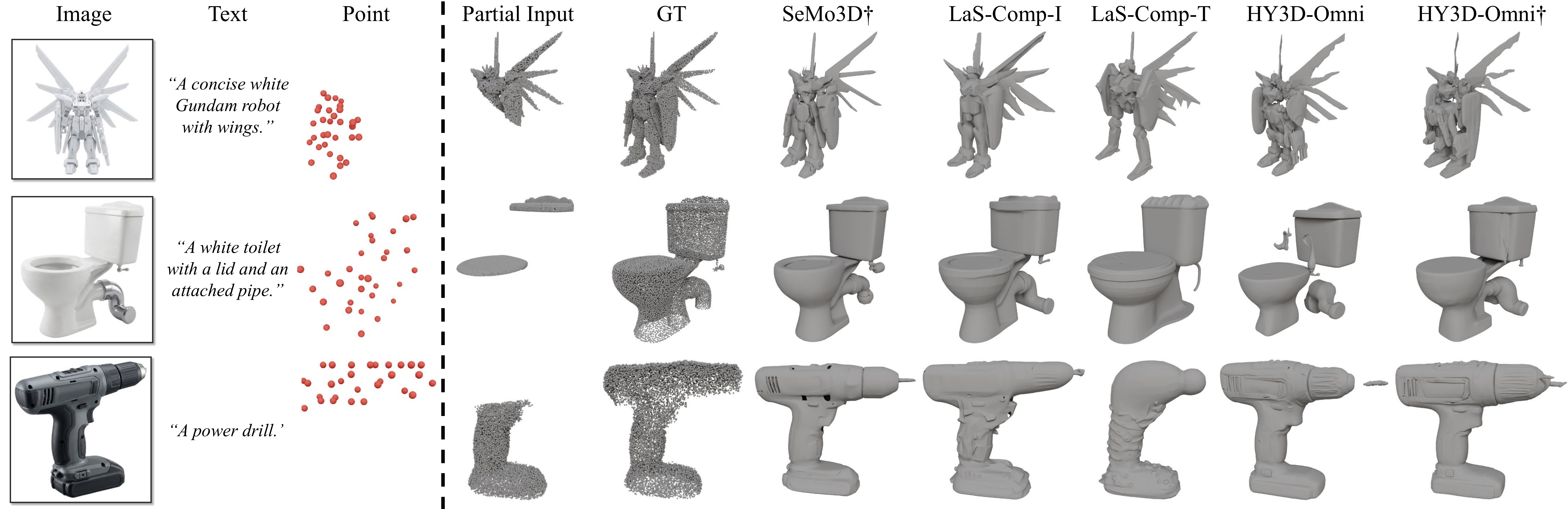}
    \vspace{-4mm}
    \caption{Qualitative comparison of 3D completion on Omni-Comp~\cite{lascomp}. Without any paired completion data, SeMo3D completes missing regions with coherent geometry that blends seamlessly with the preserved structure. $^\dagger$With additional sparse point guidance.}
    \label{fig:comp}
    \vspace{-6mm}
\end{figure*}

\subsection{Evaluation}
\label{sec:4.2}

\textbf{3D Generation.}
\cref{fig:gen} and \cref{tab:gen} present comparisons on Toys4K.
SeMo3D achieves the best LPIPS, FID, DINO, and CD, with the second-best KID, surpassing image-only, text-only, and dual-condition baselines.
\cref{fig:gen} shows why: for the cake (row 1), text-only Trellis produces the tiers and candles but misses the reference appearance; image-only methods match the appearance but miss the tiers hidden from view, while SeMo3D recovers the hidden tiers with points fixing the layout.
Images thus specify appearance, text semantics, and points spatial layout, together producing more faithful results than any single modality.

\noindent\textbf{3D Editing.}
\cref{fig:edit} and \cref{tab:edit} show editing comparisons on Edit3D-Bench.
SeMo3D leads across all three metric groups.
In region consistency (CD, PSNR, SSIM, LPIPS), the advantage stems from keeping tokens in $\mathcal{C}$ clean by construction, guaranteeing rather than approximating source fidelity: in \cref{fig:edit}, the girl's umbrella (row 2) and the man's head (row 3) remain faithful to the source in SeMo3D, whereas baselines noticeably deviate in these unedited regions.
Multi-condition guidance further improves semantic accuracy (DINO, CLIP).
SeMo3D achieves these results without paired editing data, learning editing entirely from ordinary 3D assets---a practical advantage over data-hungry methods like 3DEditFormer~\cite{3deditformer} and PartFlow~\cite{partflow}.

\begin{table}[tp]
    \caption{Quantitative results on 3D editing. Cond.\ as in \cref{tab:gen}. Edit3D-Bench~\cite{voxhammer} does not provide point conditions.}
    \vspace{-3mm}
    \label{tab:edit}
    \centering
    \renewcommand{\arraystretch}{0.8}
    \resizebox{0.99\linewidth}{!}{
        \begin{tabular}{lccccccccc}
            \toprule
            \multirow{2}{*}{Method} & \multirow{2}{*}{Cond.} & \multicolumn{4}{c}{Consistency} & \multicolumn{2}{c}{Quality} & \multicolumn{2}{c}{Accuracy} \\
            \cmidrule(lr){3-6} \cmidrule(lr){7-8} \cmidrule(lr){9-10}
            & & CD $\downarrow$ & PSNR $\uparrow$ & SSIM $\uparrow$ & LPIPS $\downarrow$ & FID $\downarrow$ & FVD $\downarrow$ & DINO $\uparrow$ & CLIP $\uparrow$ \\
            \midrule
            VoxHammer~\cite{voxhammer}         & I & 0.0111 & 31.21 & 0.973 & 0.038 & 26.35 & 586.1 & \underline{0.869} & \underline{0.281} \\
            VoxHammer~\cite{voxhammer}         & T & \underline{0.0092} & \underline{31.55} & \underline{0.976} & \underline{0.033} & \underline{23.69} & \underline{339.2} & 0.833 & 0.271 \\
            Nano3D~\cite{nano3d}               & I & 0.0101 & 31.42 & 0.975 & 0.035 & 24.12 & 388.6 & 0.861 & 0.280 \\
            3DEditFormer~\cite{3deditformer}    & I & 0.0213 & 27.44 & 0.948 & 0.061 & 28.03 & 572.4 & 0.858 & 0.278 \\
            PartFlow~\cite{partflow}            & I & 0.0153 & 30.13 & 0.967 & 0.044 & 25.33 & 437.7 & 0.858 & 0.277 \\
            \midrule
            SeMo3D (Ours)                       & I, T & \textbf{0.0088} & \textbf{37.39} & \textbf{0.986} & \textbf{0.028} & \textbf{22.46} & \textbf{113.6} & \textbf{0.875} & \textbf{0.286} \\
            \bottomrule
        \end{tabular}
    }
    \vspace{-6mm}
\end{table}
\begin{table}[tp]
    \caption{Quantitative results on 3D completion. Cond.\ as in \cref{tab:gen}. $^\dagger$With additional sparse point guidance.}
    \vspace{-3mm}
    \label{tab:comp}
    \centering
    \renewcommand{\arraystretch}{0.8}
    \resizebox{0.90\linewidth}{!}{
    \begin{tabular}{lcccc}
        \toprule
        Method & Cond. & CD $\downarrow$ & EMD $\downarrow$ & F-Score $\uparrow$ \\
        \midrule
        LaS-Comp~\cite{lascomp}             & I    & 0.0226 & 0.0375 & 0.359 \\
        LaS-Comp~\cite{lascomp}             & T    & 0.0313 & 0.0489 & 0.335 \\
        HY3D-Omni~\cite{hunyuan3d_omni}     & I, P & 0.0360 & 0.0559 & 0.347 \\
        HY3D-Omni$^\dagger$~\cite{hunyuan3d_omni}  & I, P & 0.0270 & 0.0464 & 0.412 \\
        \midrule
        SeMo3D (Ours)                        & I, T & \underline{0.0208} & \underline{0.0346} & \underline{0.468} \\
        SeMo3D$^\dagger$ (Ours)                        & I, T, P & \textbf{0.0133} & \textbf{0.0223} & \textbf{0.541} \\
        \bottomrule
    \end{tabular}
    }
    \vspace{-8mm}
\end{table}

\noindent\textbf{3D Completion.}
\cref{fig:comp} and \cref{tab:comp} report completion results on Omni-Comp.
SeMo3D achieves the best CD, EMD, and F-Score.
Even without additional point guidance, SeMo3D outperforms all compared methods across all three completion metrics.
Unlike point-guided HY3D-Omni, which regenerates whole shapes, SeMo3D preserves existing geometry, synthesizing only the missing part, despite its irregular, non-box shape.
In \cref{fig:comp}, completed regions blend seamlessly with the existing structure.
No paired completion data is used; SeMo3D derives this capability purely from region-selective denoising on standard 3D assets, demonstrating the generality of selective modeling.

\subsection{Ablation Study}
\label{sec:4.3}

\begin{table*}[tp]
    \caption{Quantitative ablation. (i)--(iv): condition combinations; (v)--(vii): full-model component ablations. Editing consistency metrics are omitted given identical context tokens across variants. Edit3D-Bench lacks points, so (iii)--(iv) omit editing and (v)--(vii) edit with I,\,T.}
    \vspace{-3mm}
    \label{tab:ablation}
    \centering
    \renewcommand{\arraystretch}{0.8}
    \resizebox{0.90\linewidth}{!}{
    \begin{tabular}{clccccccccccccc}
        \toprule
        & \multirow{2}{*}{Configuration} & \multirow{2}{*}{Cond.} & \multicolumn{5}{c}{Generation} & \multicolumn{4}{c}{Editing} & \multicolumn{3}{c}{Completion} \\
        \cmidrule(lr){4-8} \cmidrule(lr){9-12} \cmidrule(lr){13-15}
        & & & LPIPS $\downarrow$ & FID $\downarrow$ & KID ($\times$100) $\downarrow$ & DINO $\uparrow$ & CD $\downarrow$ & FID $\downarrow$ & FVD $\downarrow$ & DINO $\uparrow$ & CLIP $\uparrow$ & CD $\downarrow$ & EMD $\downarrow$ & F-Score $\uparrow$ \\
        \midrule
        (i) & SeMo3D (Ours) & I       & 0.113 & 5.49 & 0.074 & 0.904 & 0.0334 & 22.53 & 118.2 & 0.873 & 0.284 & 0.0207 & 0.0346 & 0.471 \\
        (ii) & SeMo3D (Ours) & I, T    & 0.112 & 5.34 & 0.069 & 0.911 & 0.0326 & 22.46 & 113.6 & 0.875 & 0.286 & 0.0208 & 0.0346 & 0.468 \\
        (iii) & SeMo3D (Ours) & I, P    & 0.082 & 5.06 & 0.065 & 0.926 & 0.0116 & -- & -- & -- & -- & 0.0134 & 0.0222 & 0.542 \\
        (iv) & SeMo3D (Ours) & I, T, P & 0.079 & 4.99 & 0.059 & 0.926 & 0.0118 & -- & -- & -- & -- & 0.0133 & 0.0223 & 0.541 \\
        \midrule
        (v) & w/o Role Emb.      & I, T, P & 0.078 & 5.01 & 0.057 & 0.925 & 0.0118 & 22.52 & 114.2 & 0.866 & 0.278 & 0.0156 & 0.0249 & 0.528 \\
        (vi) & Point via Cross-Attn.  & I, T, P & 0.102 & 5.32 & 0.066 & 0.917 & 0.0244 & 22.47 & 112.9 & 0.875 & 0.285 & 0.0178 & 0.0276 & 0.499 \\
        (vii) & Standard CFG Dropout  & I, T, P & 0.110 & 5.44 & 0.072 & 0.902 & 0.0331 & 22.51 & 116.9 & 0.872 & 0.283 & 0.0201 & 0.0350 & 0.467 \\
        \bottomrule
    \end{tabular}
    }
    \vspace{-1mm}
\end{table*}
\begin{figure*}[tp]
    \centering
    \includegraphics[width=0.90\linewidth]{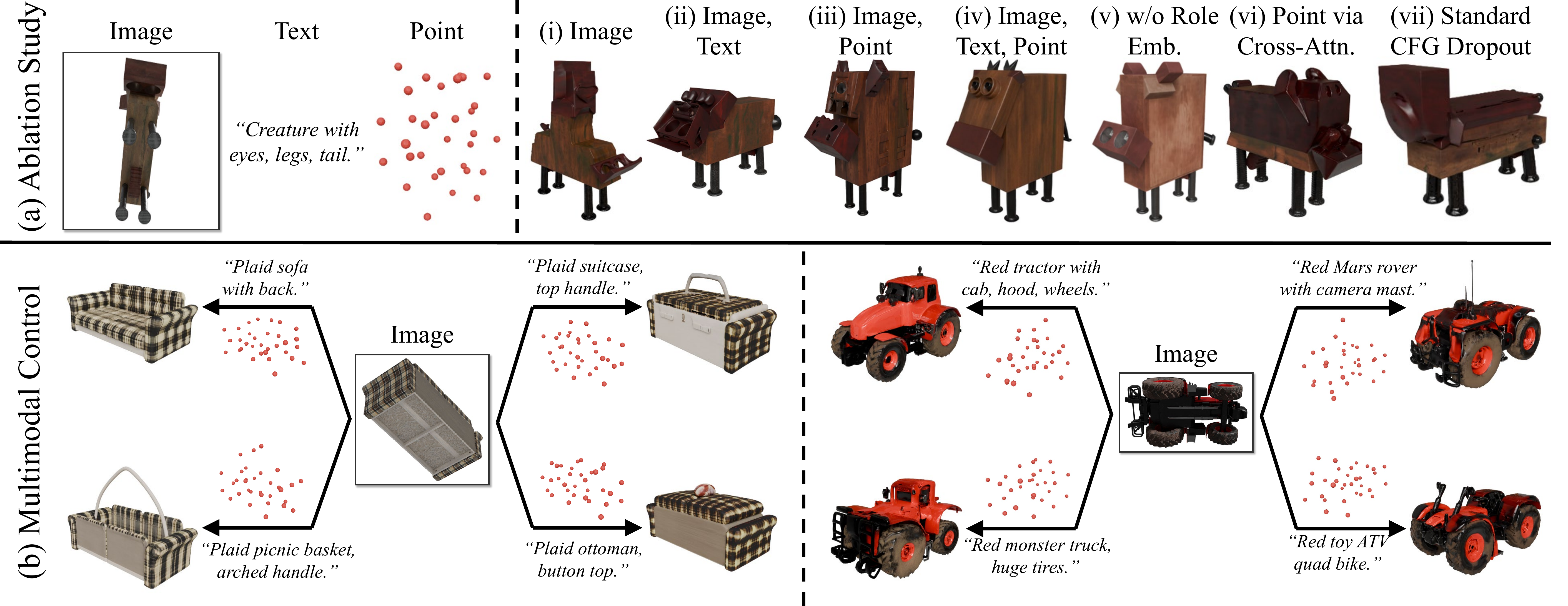}
    \vspace{-4mm}
    \caption{(a)~Qualitative ablation. (b)~Multimodal control: with a fixed image, varying text and points yields diverse yet faithful results.}
    \label{fig:ablation}
    \vspace{-6mm}
\end{figure*}
\begin{figure}[tp]
    \centering
    \includegraphics[width=0.90\linewidth]{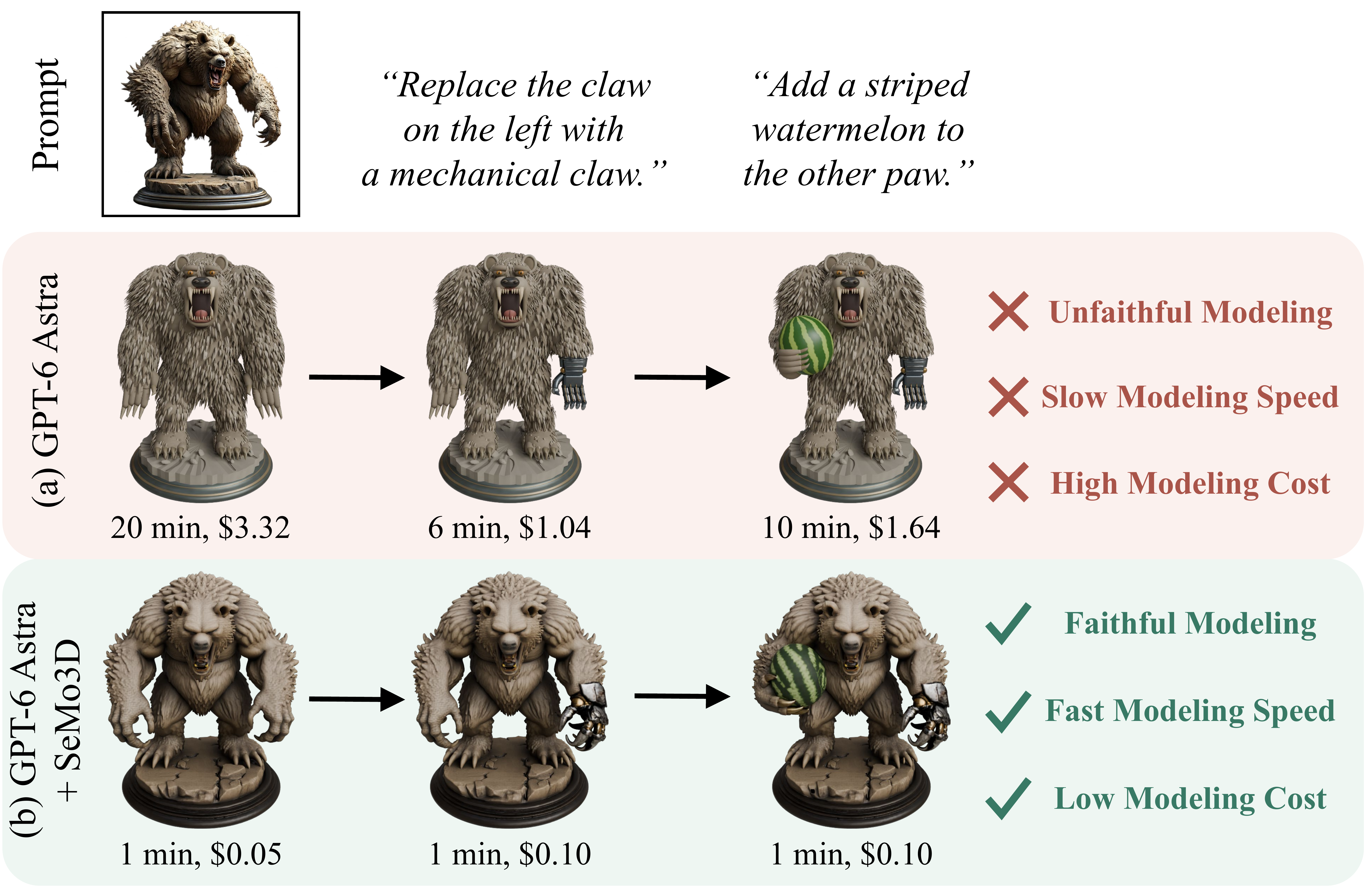}
    \vspace{-3mm}
    \caption{Agentic 3D modeling with GPT-6 Astra~\cite{chatgpt} and Codex: (a)~direct Blender coding~\cite{blender}; (b)~SeMo3D integration, yielding higher modeling fidelity, faster execution, and lower agent cost.}
    \label{fig:agent}
    \vspace{-5mm}
\end{figure}

\textbf{Multi-Conditioning.}
We ablate the optional text and point conditions in \cref{tab:ablation}~(i)--(iv) and \cref{fig:ablation}(a).
Starting from image-only input~(i), adding text~(ii) improves perceptual quality and semantic fidelity (FID, DINO) in generation and editing but barely changes geometry: the creature gains eyes and a tail, yet its body stays a flat box.
Adding points instead~(iii) yields larger gains in geometry (CD, EMD, F-Score): the creature's body better matches the point layout.
The full model~(iv) achieves the best perceptual quality and semantic fidelity while matching~(iii) in geometry, combining visual, semantic, and spatial cues no single modality provides.
\cref{fig:ablation}(b) shows that, with a fixed image, varying text and points yields a sofa, suitcase, basket, or ottoman in one case and a tractor, rover, truck, or ATV in the other, demonstrating fine-grained multimodal control.
\begin{figure}[tp]
    \centering
    \includegraphics[width=0.90\linewidth]{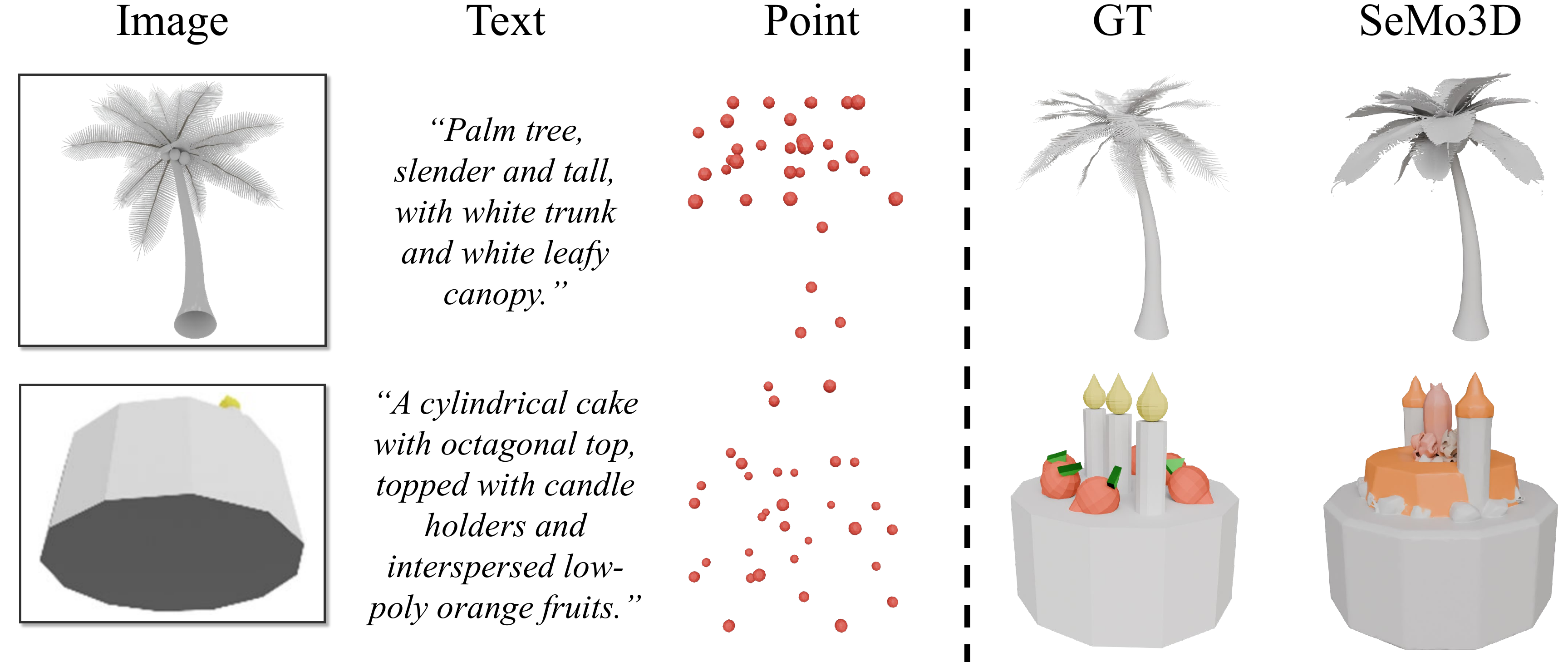}
    \vspace{-3mm}
    \caption{Failure cases: coarse fronds and missing fruits.}
    \label{fig:failure}
    \vspace{-5mm}
\end{figure}
\textbf{Role Embedding.}
Removing the role embedding~(v) barely affects generation, where all tokens are selected, but degrades editing and completion (\cref{tab:ablation}), as distinguishing clean context from noisy tokens becomes harder.
\textbf{Point Embedding vs.\ Cross-Attention.}
Injecting points via cross-attention~(vi) doubles the generation CD and degrades completion (\cref{tab:ablation}); the creature in \cref{fig:ablation}(a) elongates, ignoring the point layout, which confirms that self-attention integrates points more effectively.
\textbf{Compositional Condition Dropout.}
Replacing compositional condition dropout with standard Classifier-Free Guidance~\cite{cfg} (CFG) dropout~(vii), which retains either all available conditions or none, weakens text and point guidance: metrics nearly revert to image-only~(i), and the output in \cref{fig:ablation}(a) has an elongated body, disregarding text and points.
\noindent\textbf{SeMo3D Integration.}
Compared with direct Blender coding~\cite{blender}, equipping the same agent with SeMo3D improves modeling fidelity while reducing execution time and agent cost (\cref{fig:agent}).

\section{Conclusion}
\label{sec:5}

We present SeMo3D, a unified framework that casts 3D generation, editing, and completion as selective modeling---synthesizing only the designated region while preserving the rest.
Around this core idea, we design region-selective denoising and Omni3DiT, enabling one model to handle all three tasks without task-specific paired data while accepting complementary image, text, and point guidance.
Experiments show that SeMo3D surpasses specialized state-of-the-art methods across all three tasks.
Agent integration further enables an intuitive, efficient 3D creation workflow.

\noindent\textbf{Limitations and Future Work.}
SeMo3D inherits Trellis2's structured latents~\cite{trellis2}, whose limited resolution cannot capture very fine structures such as thin palm fronds (\cref{fig:failure}, top); higher-fidelity 3D representations~\cite{anysurf,p2voxel,supervoxelgpt} could alleviate this.
Moreover, the CLIP~\cite{clip} text encoder struggles with complex compositional prompts: the orange fruits are reduced to orange texture rather than generated as objects (\cref{fig:failure}, bottom); stronger vision-language encoders~\cite{qwen3,qwen3_vl} are a promising remedy.

\section{LLM Usage Statement}
\label{sec:6}

Codex was used throughout the manuscript for language polishing, concision, and grammar checking. It was also used to obtain reviewer-style feedback, assist with code development, organize experimental results, and check references.
GPT-Image was used to generate and edit reference images for the agentic 3D modeling experiments.
In these experiments, GPT-6 Astra with Codex served as the agent, preparing inputs, determining edit regions, invoking SeMo3D, and generating code for the direct Blender baseline.
The authors reviewed all AI-assisted content and take full responsibility for the scientific content, code, experimental results, references, and conclusions.

{
    \small
    \bibliographystyle{ieeenat_fullname}
    \bibliography{main}
}

\maketitlesupplementary

This supplementary material provides training details (\cref{sec:supp_training}), experimental details (\cref{sec:supp_experiment}), robustness analysis (\cref{sec:supp_robustness}), and additional results (\cref{sec:supp_results}).

\section{Training Details}
\label{sec:supp_training}

\noindent\textbf{Training Data.}
We train on the training split of Trellis-500K~\cite{trellis}, filtering out assets with an aesthetic score below 4.5.
Each sample pairs one asset with a randomly chosen rendered view as the image condition and a randomly chosen dataset caption as the text condition.
No paired editing or completion data is used.

\noindent\textbf{Point Conditions.}
Points lie on the $16^3$ SS grid, with at most 32 per sample.
During training, we first use farthest point sampling (FPS) to obtain 32 spatially distributed points from occupied voxels inside $\mathcal{S}$, then draw $k$ uniformly from 1 to 32 and randomly select $k$ of these points; unused slots are filled with masked padding tokens.
This random subsampling improves robustness to variations in the number and spatial distribution of user-provided point conditions (\cref{sec:supp_robustness}).

\noindent\textbf{Hyperparameters.}
All three stages are initialized from Trellis2~\cite{trellis2}, with settings listed in \cref{tab:supp_hparams}.
We use AdamW with a learning rate of $5{\times}10^{-5}$, weight decay of 0.01, an EMA rate of 0.9999, and bfloat16 mixed precision.

\noindent\textbf{Region-Selective Denoising.}
Full-region and local-region modes are sampled with equal probability.
In the local-region mode, we sample a 3D box with random side lengths (\cref{tab:supp_hparams}) and position, restricted to the half of the asset facing the condition camera so that the selected region is observed by the image.
Sparse boxes are rejected, and accepted ones are shrunk to fit their occupied voxels.
The SS grid has resolution $N{=}16$; at inference, its partition is transferred to the $32^3$ SLat grid by nearest-neighbor upsampling.

\noindent\textbf{Compositional Condition Dropout.}
Each step keeps one condition combination and nulls the rest.
For the SS stage, \{I, T, P\}, \{I, T\}, and \{I, P\} are each sampled with probability 0.2, and \{I\}, \{T\}, \{P\}, and $\emptyset$ with 0.1 each.
For the SLat stages, \{I, T\}, \{I\}, \{T\}, and $\emptyset$ are sampled with probabilities 0.5, 0.2, 0.2, and 0.1, respectively.
The image is thus withheld in 30\% of steps, and $\emptyset$ trains the unconditional branch for classifier-free guidance~\cite{cfg}.
In the standard CFG dropout ablation, each stage retains all available conditions with probability 0.9 and drops them all with probability 0.1: the SS stage uses only \{I, T, P\} or $\emptyset$, and the SLat stages use only \{I, T\} or $\emptyset$.

\section{Experimental Details}
\label{sec:supp_experiment}

\subsection{Benchmarks and Metrics}
\label{sec:supp_exp_metrics}

\noindent\textbf{Generation.}
We use the Toys4K~\cite{toys4k} test split of Trellis-500K, which is disjoint from the training split, keeping the 3,180 objects with text annotations (105 categories).
Each object comes with one condition image and one caption, and its point condition consists of 32 points sampled by FPS from the asset's occupancy.
Generated and ground-truth meshes are rendered from 8 views at $15^\circ$ elevation with $45^\circ$ azimuth spacing.
LPIPS~\cite{lpips} and DINOv2~\cite{dinov2} similarity are averaged over view-aligned render pairs, while FID~\cite{fid} and KID~\cite{kid} compare the two render sets using Inception-v3 features~\cite{inv3}.
CD uses 10K surface points per mesh, with each mesh normalized to a unit box.

\begin{table}[tp]
    \caption{Per-stage training settings.}
    \label{tab:supp_hparams}
    \centering
    \renewcommand{\arraystretch}{0.9}
    \resizebox{0.95\linewidth}{!}{
    \begin{tabular}{lccc}
        \toprule
        & SS & SLat (shape) & SLat (texture) \\
        \midrule
        Grid resolution & $16^3$ & $32^3$ & $32^3$ \\
        Width / blocks / heads & \multicolumn{3}{c}{1536 / 30 / 12} \\
        Conditions & I, T, P & I, T & I, T \\
        Training steps & 200K & 100K & 100K \\
        Batch size & 32 & 32 & 32 \\
        Box side length (voxels) & 4--16 & 8--32 & 8--32 \\
        \bottomrule
    \end{tabular}
    }
\end{table}

\noindent\textbf{Editing.}
Edit3D-Bench~\cite{voxhammer} contains 300 editing cases, each with a source asset, a 3D edit region, an edited image, and a text prompt.
Edited and source assets are rendered from 16 views at $15^\circ$ and $30^\circ$ elevations.
Consistency is measured outside the edit region: PSNR, SSIM~\cite{ssim}, and LPIPS on unmasked pixels of view-aligned renders, and CD on 20K surface points with those inside the edit region removed.
Quality is measured by FID between source and edited renders and by FVD between their 120-frame orbit videos.
Accuracy is measured on 6 frontal views by DINOv2 similarity to the edited image and CLIP~\cite{clip} similarity to the prompt.

\begin{figure}[tp]
    \centering
    \includegraphics[width=\linewidth]{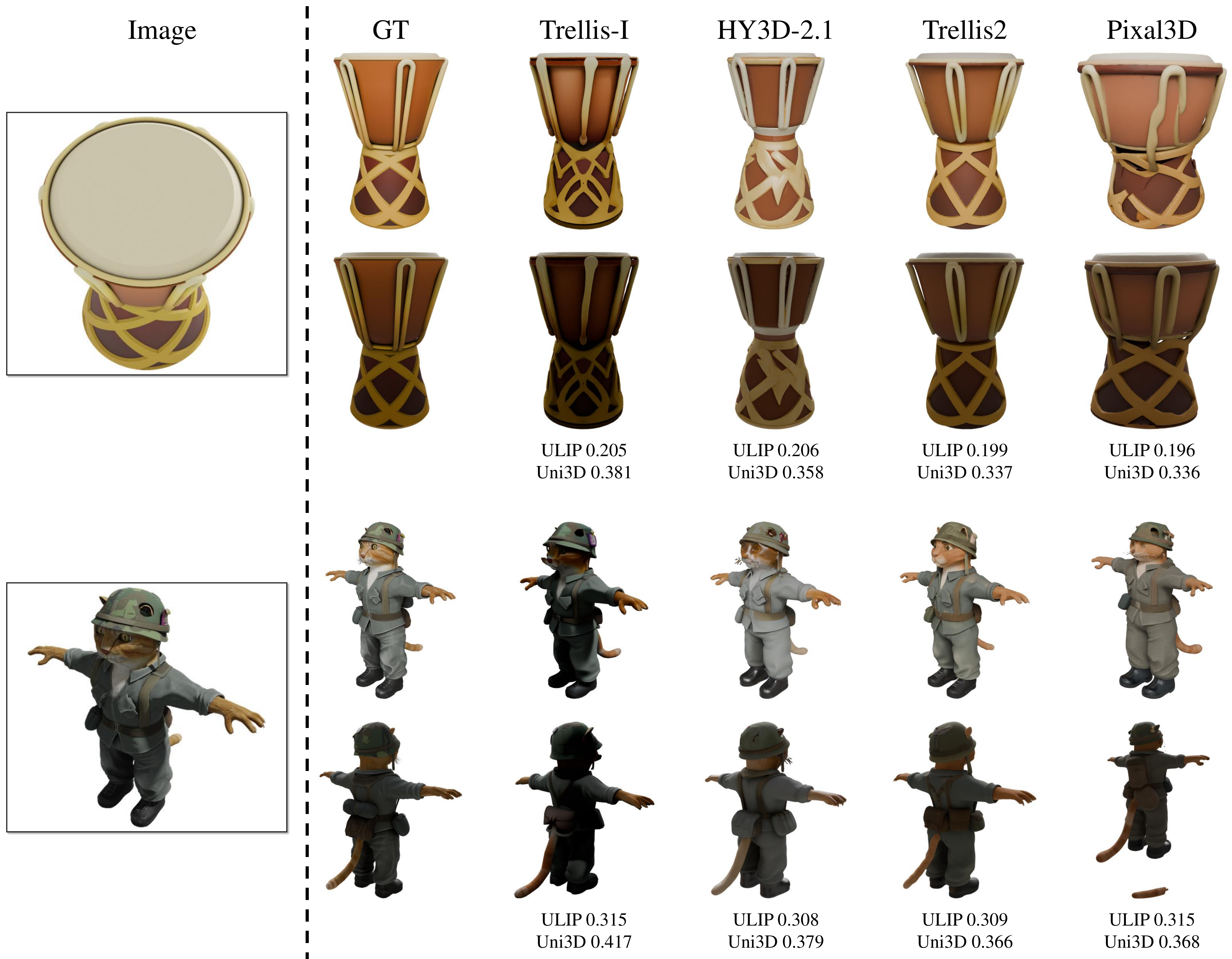}
    \caption{ULIP and Uni3D scores disagree with visual quality: Trellis2 is closer to the ground truth than Trellis-I but scores lower.}
    \label{fig:supp_uni3d}
\end{figure}

\begin{figure}[tp]
    \centering
    \includegraphics[width=\linewidth]{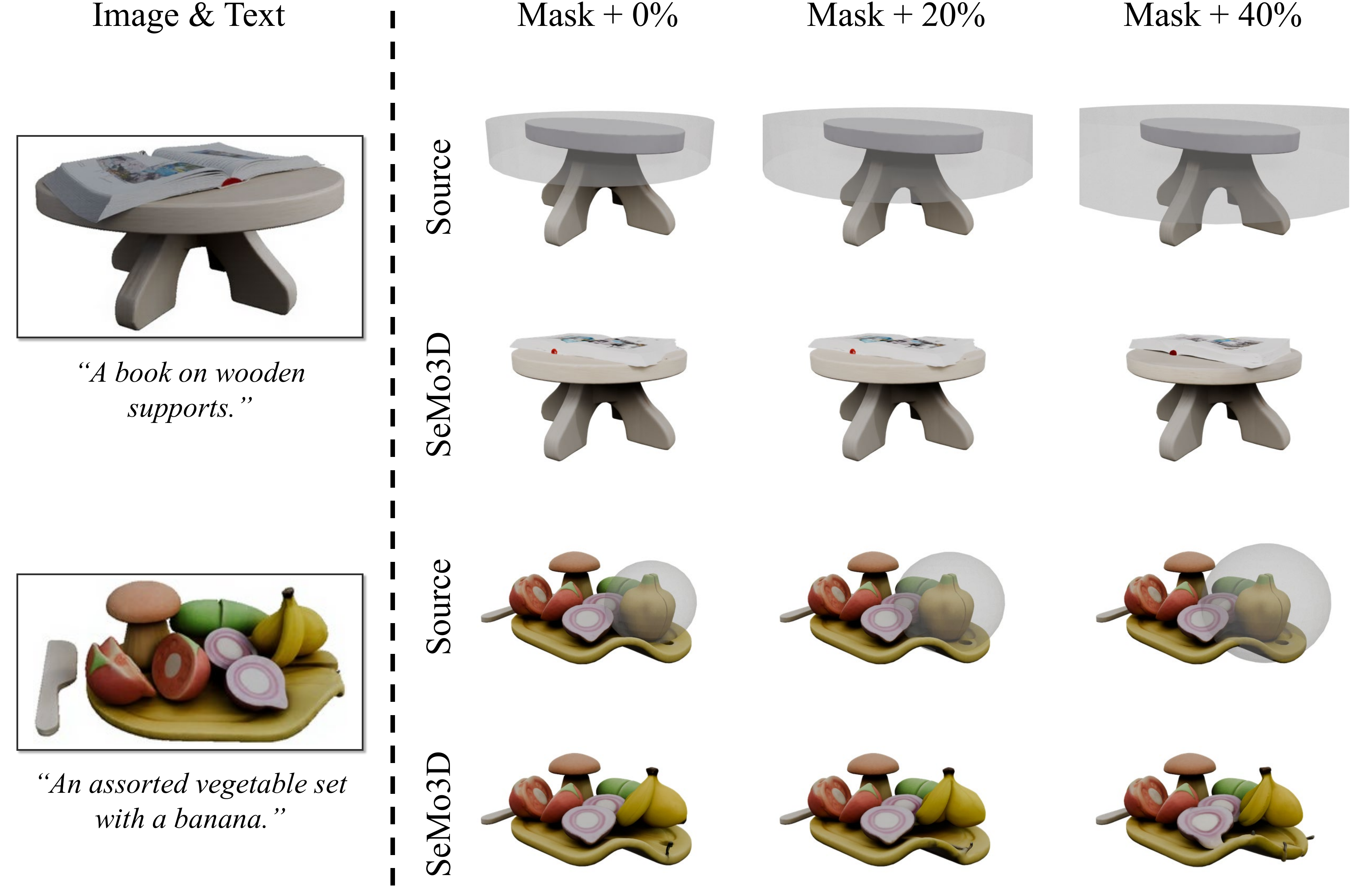}
    \caption{Robustness to mask size. SeMo3D yields consistent edits as the edit region (translucent) grows by up to 40\%.}
    \label{fig:supp_mask}
\end{figure}

\noindent\textbf{Completion.}
Omni-Comp~\cite{lascomp} contains 180 partial inputs created by single-view scanning, part removal, and random cropping.
For SeMo3D, the point condition consists of 32 points sampled by FPS from ground-truth surface points farther than 0.02 from the partial input.
The no-point variant uses the same image and text conditions without these additional points.
We sample 16K points from both the completed mesh and the ground truth, and apply the same normalization transform to both so that they remain aligned in a shared coordinate system.
CD, EMD, and F-Score at a 0.01 threshold are computed on the whole shape, reflecting both the completed region and the preserved input.

\noindent\textbf{Why Not 3D Embedding Metrics?}
ULIP~\cite{ulip} and Uni3D~\cite{uni3d} scores do not reliably reflect generation quality.
As shown in \cref{fig:supp_uni3d}, Trellis2 is visibly closer to the ground truth than Trellis-I and outperforms it in Tab.~1 of the main paper, yet receives lower scores.
We therefore rely on metrics computed against the ground truth.

\begin{figure}[tp]
    \centering
    \includegraphics[width=\linewidth]{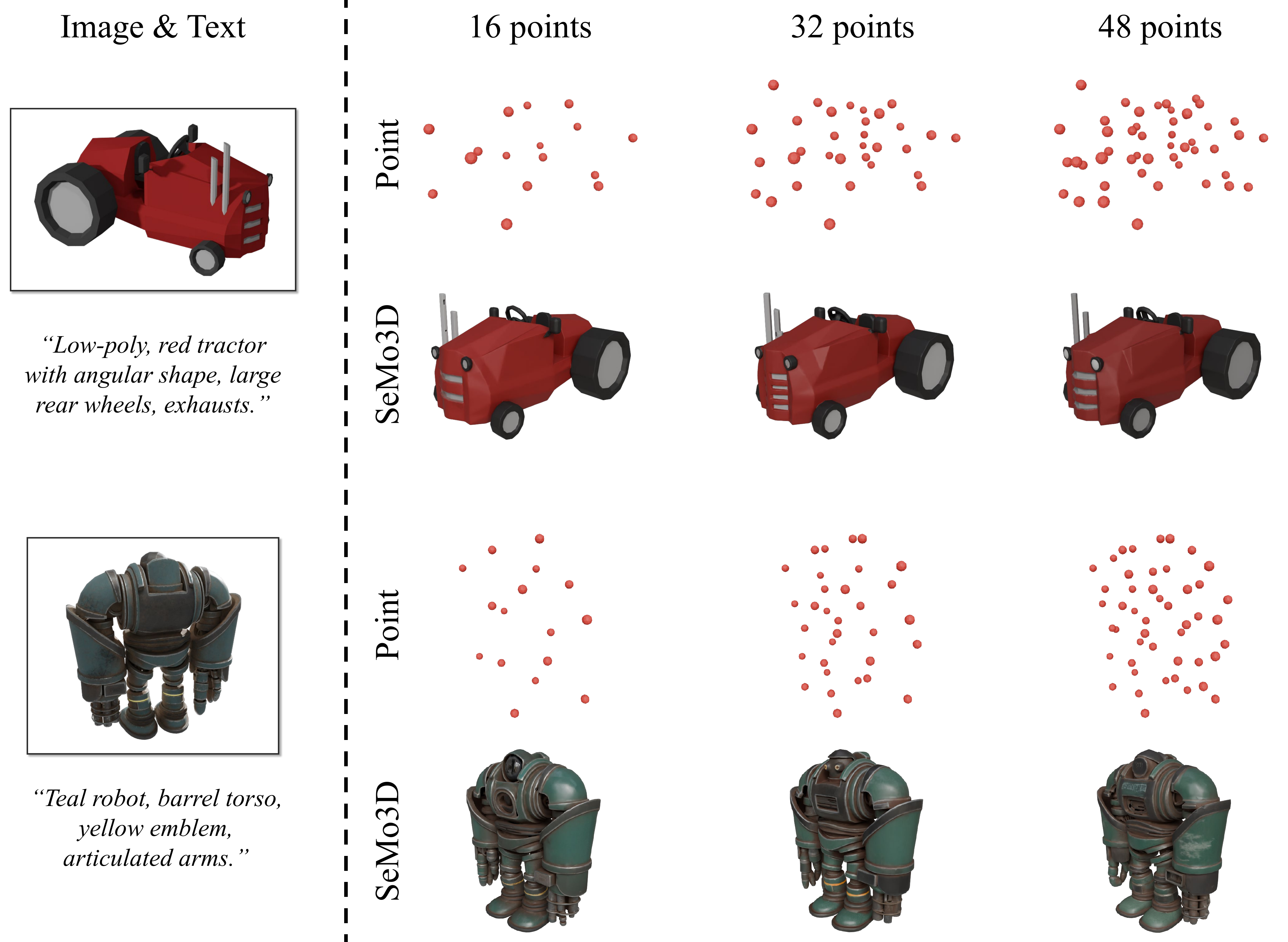}
    \caption{Robustness to the number of condition points. With the image and text fixed, SeMo3D preserves the overall geometry across 16, 32, and 48 points, despite being trained with at most 32.}
    \label{fig:supp_point_num}
\end{figure}

\begin{figure}[tp]
    \centering
    \includegraphics[width=\linewidth]{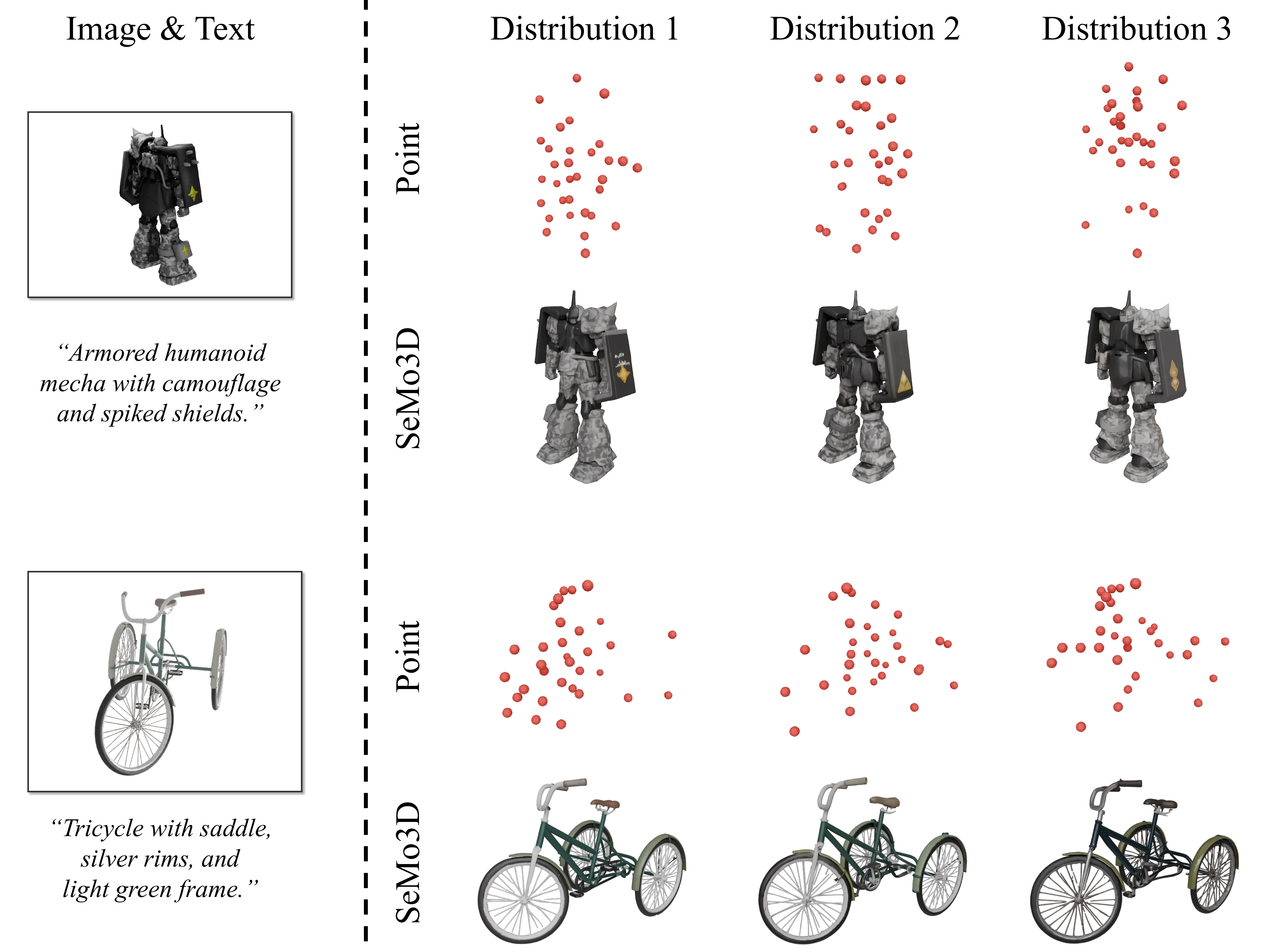}
    \caption{Robustness to the spatial distribution of condition points. With the image and text fixed, SeMo3D produces consistent overall structure across three point distributions, with variations in local details.}
    \label{fig:supp_point_dist}
\end{figure}

\subsection{Baselines and Inference}
\label{sec:supp_exp_baselines}

\noindent\textbf{Baselines.}
All baselines use their official settings and receive every condition they support, with variable-resolution outputs standardized to 512.
For editing, the training-free VoxHammer~\cite{voxhammer} and Nano3D~\cite{nano3d} take the same ground-truth edit region as SeMo3D, whereas the feed-forward 3DEditFormer~\cite{3deditformer} and PartFlow~\cite{partflow} do not accept masks.
For completion, HY3D-Omni~\cite{hunyuan3d_omni} takes the partial input as its point condition, and HY3D-Omni$^\dagger$ additionally receives sparse point guidance.

\noindent\textbf{SeMo3D Inference.}
We use an Euler flow sampler with classifier-free guidance.
For generation, all stages use 12 steps with guidance scales of 7.5, 7.5, and 1.0 for the SS, shape, and texture stages, respectively.
For editing and completion, the SS stage uses 50 steps and the SLat stages use 12, all with a guidance scale of 3.0.

\subsection{Agentic Interactive 3D Creation}
\label{sec:supp_exp_agent}

We use GPT-6 Astra~\cite{chatgpt} with Codex as the agent.
Each sequence starts with an image or a partial point cloud, followed by editing requests.

\noindent\textbf{Generation.}
The agent directly invokes SeMo3D's generation mode with the input image to obtain a 3D asset.

\noindent\textbf{Completion.}
The agent first renders a depth map of the partial point cloud and uses GPT-Image to infer an RGB image of the complete asset.
It then invokes SeMo3D's completion mode with this image and the partial point cloud to recover the full asset.

\noindent\textbf{Editing.}
Given an editing instruction, the agent uses GPT-Image to produce an edited image and autonomously determines the target region.
It then invokes SeMo3D's editing mode with the edited image, the target-region mask, and the source asset's cached latent representation to obtain the edited asset.
Each output's latent representation is cached for subsequent edits.

\noindent\textbf{Direct-Coding Baseline.}
The same agent creates and modifies assets entirely by writing Blender~\cite{blender} code, without using SeMo3D.

\section{Robustness}
\label{sec:supp_robustness}

\noindent\textbf{Mask Size.}
User-specified masks are rarely tight, so we enlarge each Edit3D-Bench edit region by up to 40\% about its center.
As shown in \cref{fig:supp_mask}, SeMo3D yields nearly identical edits across all sizes, without adding spurious geometry.
Since the SS stage decides occupancy within $\mathcal{S}$, a looser mask does not force content into it.

\noindent\textbf{Number of Condition Points.}
We vary the number of condition points across 16, 32, and 48, while keeping the image and text conditions fixed.
As shown in \cref{fig:supp_point_num}, the generated assets retain consistent overall geometry across all three counts.
The 48-point results further show generalization beyond the training maximum of 32 points.

\noindent\textbf{Spatial Distribution of Condition Points.}
We also compare three spatial distributions of condition points for each asset, keeping the image and text conditions fixed.
As shown in \cref{fig:supp_point_dist}, SeMo3D preserves the overall structure of the mecha and tricycle across these distributions, while local geometry and appearance may vary.
Together, these results indicate robustness to variations in both the number and spatial distribution of point conditions.

\section{Additional Results}
\label{sec:supp_results}
\begin{figure*}[tp]
    \centering
    \includegraphics[width=0.75\linewidth]{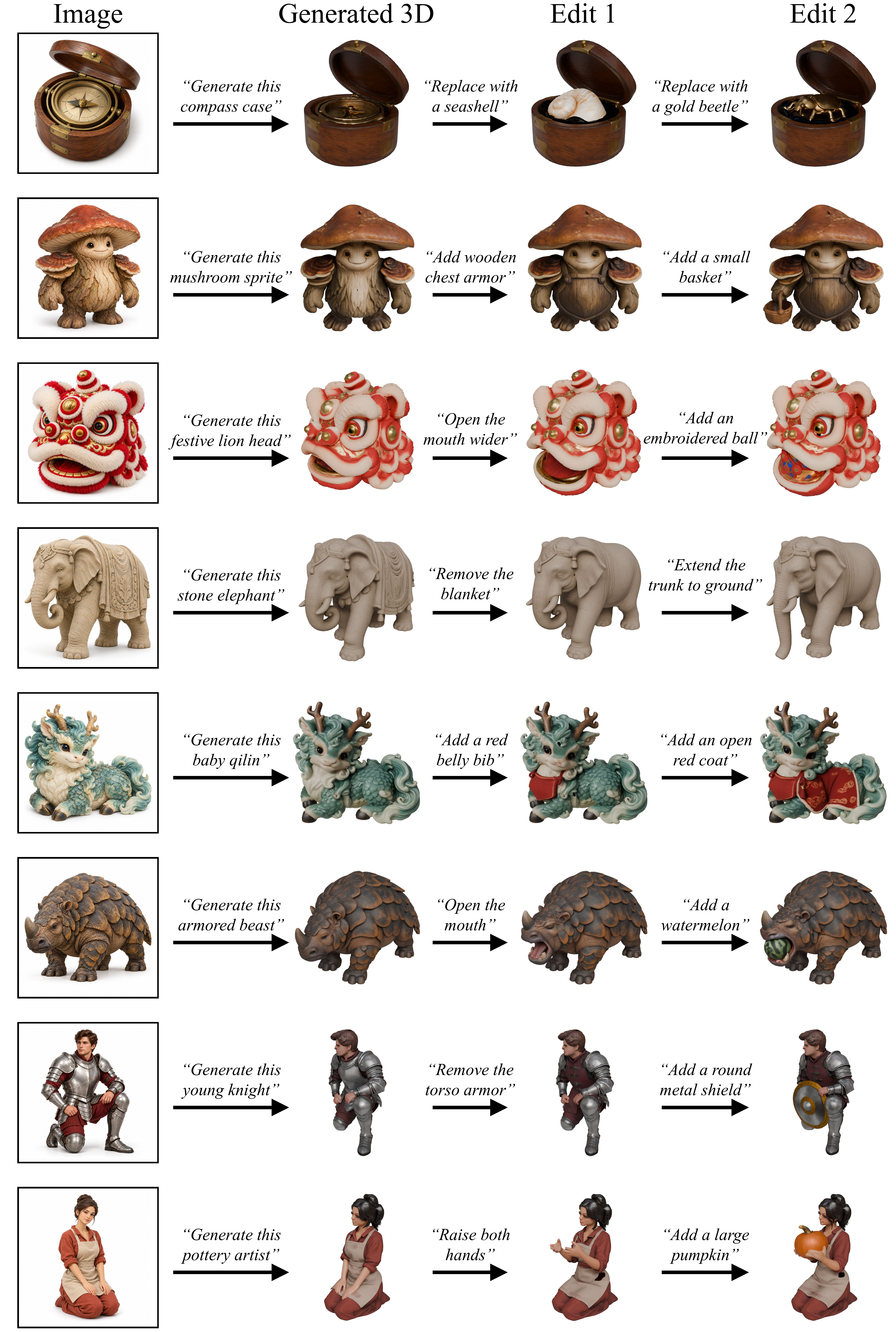}
    \caption{Additional agentic interactive 3D modeling: generation followed by two edits.}
    \label{fig:supp_agent_show1}
\end{figure*}

\begin{figure*}[tp]
    \centering
    \includegraphics[width=0.75\linewidth]{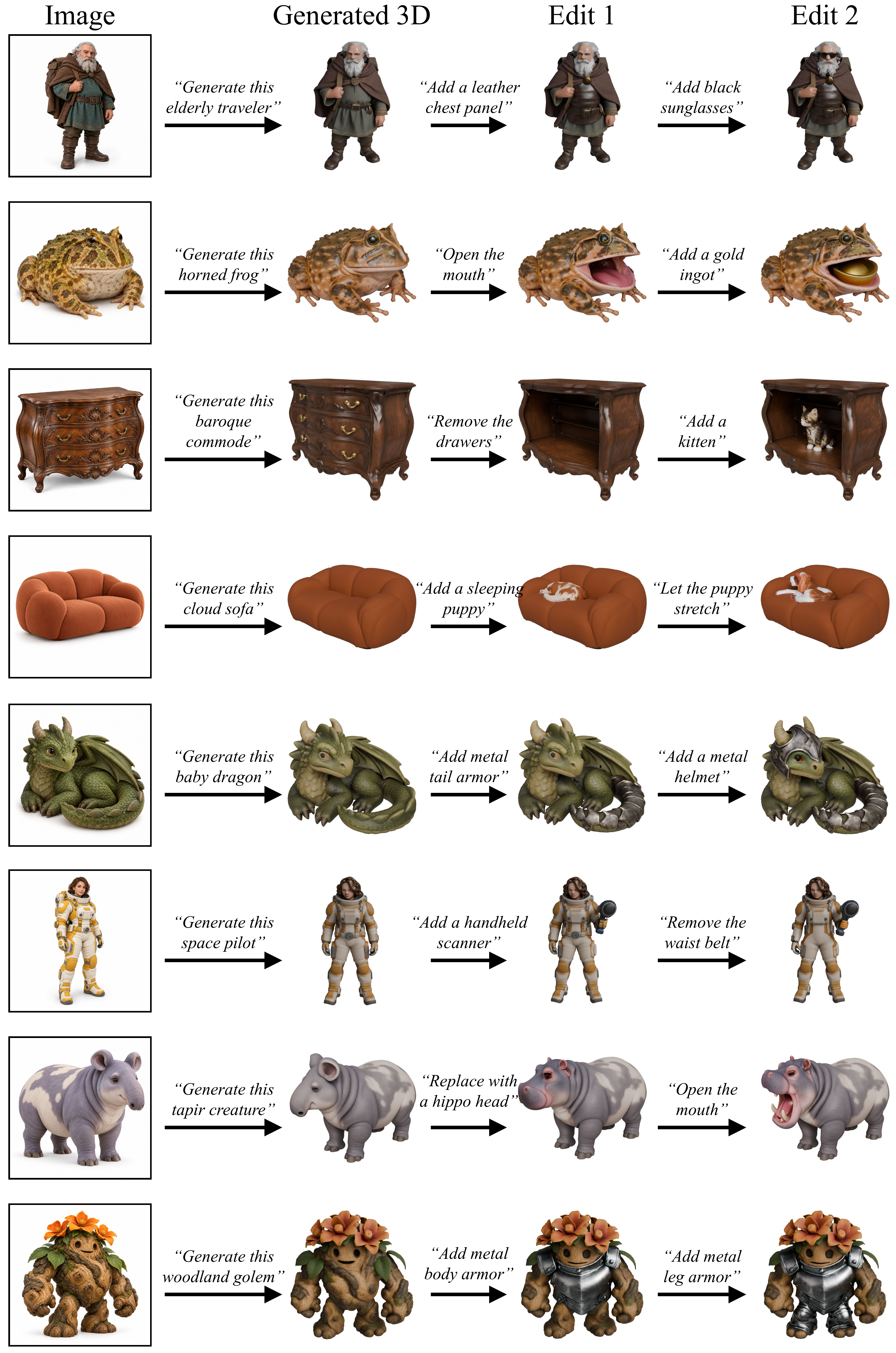}
    \caption{Additional agentic interactive 3D modeling: generation followed by two edits.}
    \label{fig:supp_agent_show2}
\end{figure*}

\begin{figure*}[tp]
    \centering
    \includegraphics[width=0.75\linewidth]{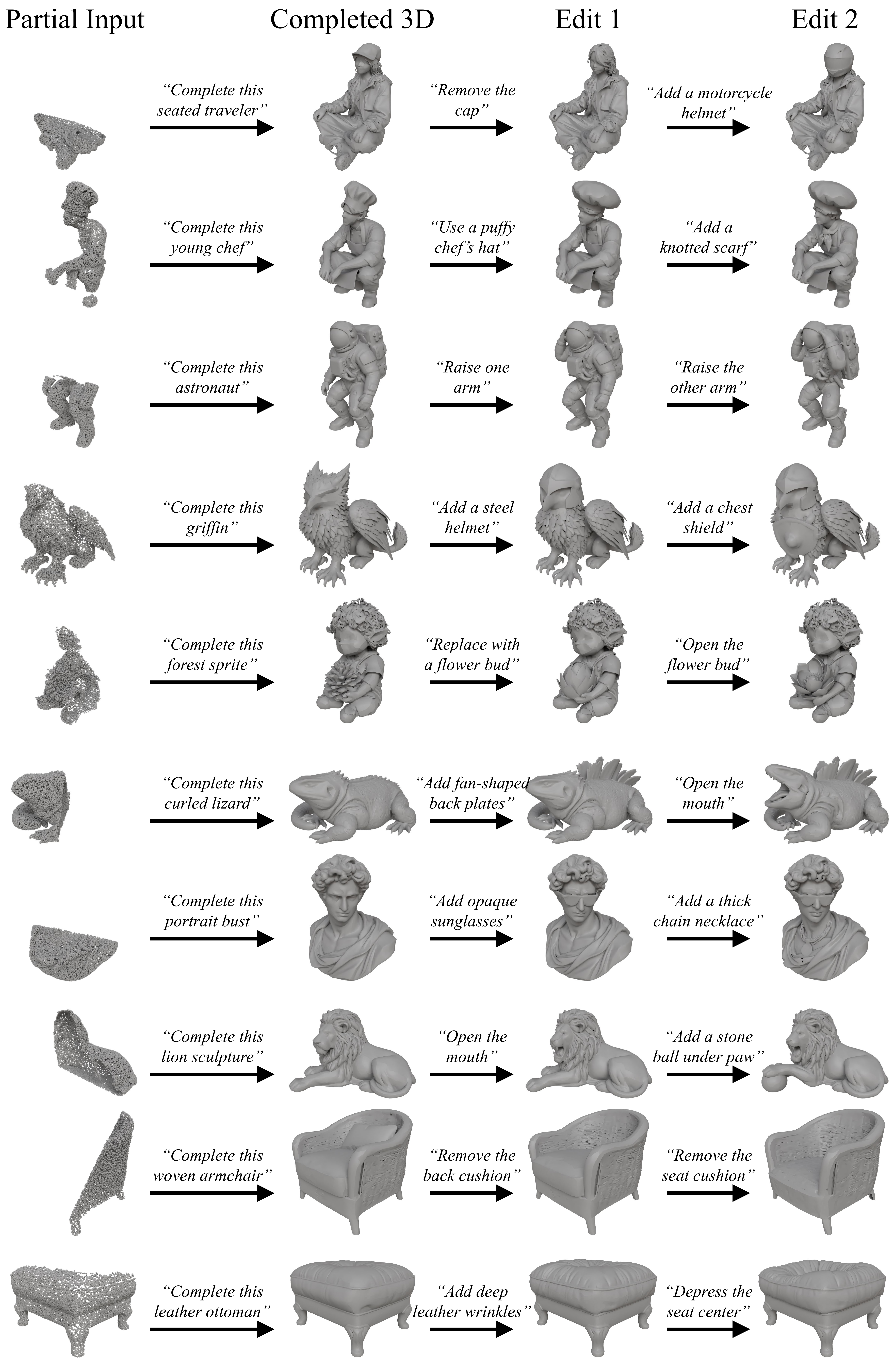}
    \caption{Additional agentic interactive 3D modeling: completion followed by two edits.}
    \label{fig:supp_agent_show3}
\end{figure*}

\begin{figure*}[tp]
    \centering
    \includegraphics[width=0.75\linewidth]{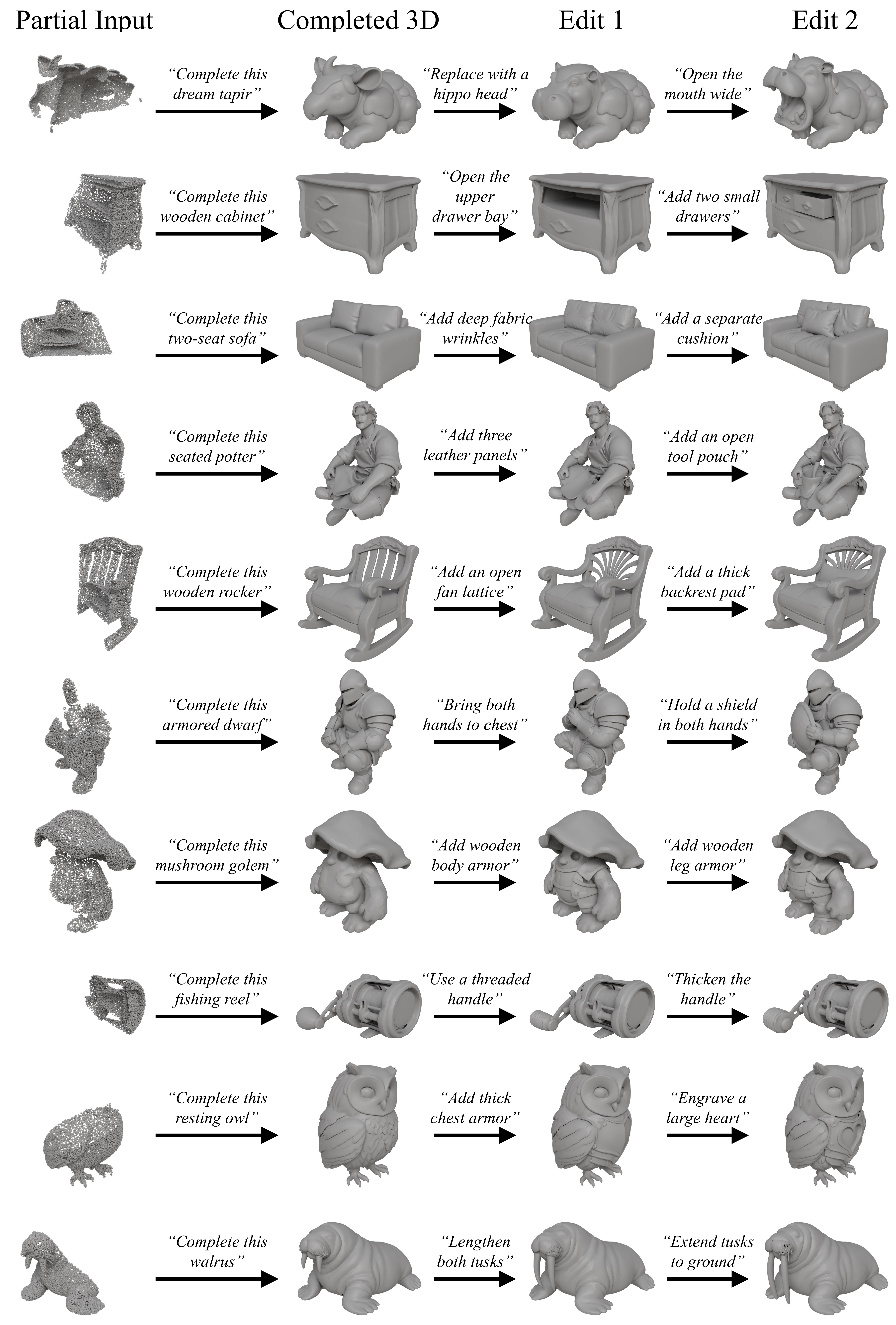}
    \caption{Additional agentic interactive 3D modeling: completion followed by two edits.}
    \label{fig:supp_agent_show4}
\end{figure*}

\textbf{Agentic Interactive 3D Modeling.}
Using the setup in \cref{sec:supp_exp_agent}, \cref{fig:supp_agent_show1,fig:supp_agent_show2} show generation followed by two edits, while \cref{fig:supp_agent_show3,fig:supp_agent_show4} start with completion.
The examples span human characters, animals, fantasy creatures, furniture, and everyday objects, covering accessory addition, part removal and replacement, pose adjustment, and local geometric refinement.
Tasks range from replacing the contents of a compass case and adding armor to a creature to modifying the backrest of a completed chair.
These results illustrate SeMo3D's versatility across asset categories and its ability to support successive user-directed changes while maintaining consistency in unedited regions.

\begin{figure*}[tp]
    \centering
    \includegraphics[width=0.85\linewidth]{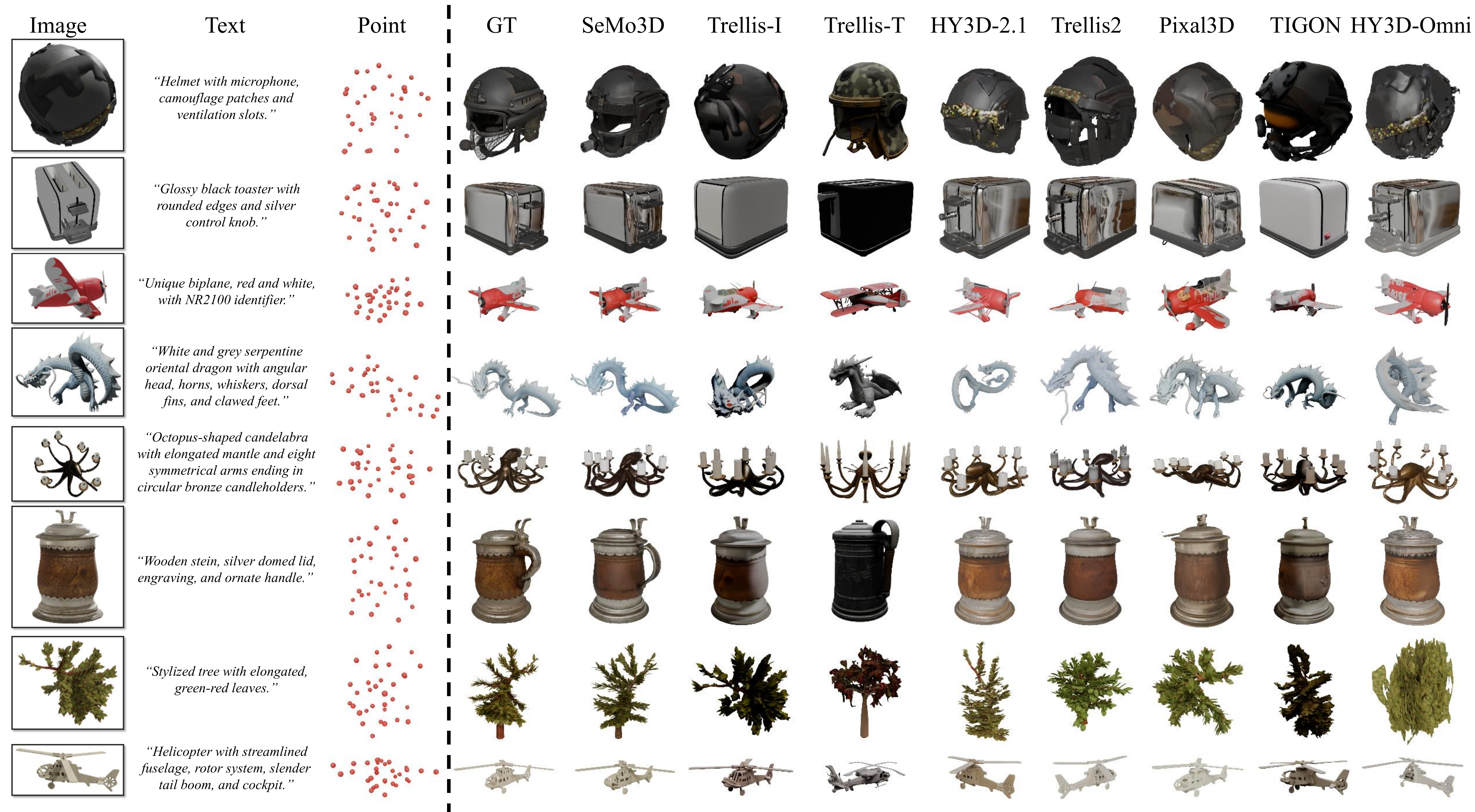}
    \caption{Additional qualitative comparison of 3D generation on Toys4K~\cite{toys4k}.}
    \label{fig:supp_gen}
\end{figure*}

\begin{figure*}[tp]
    \centering
    \includegraphics[width=0.85\linewidth]{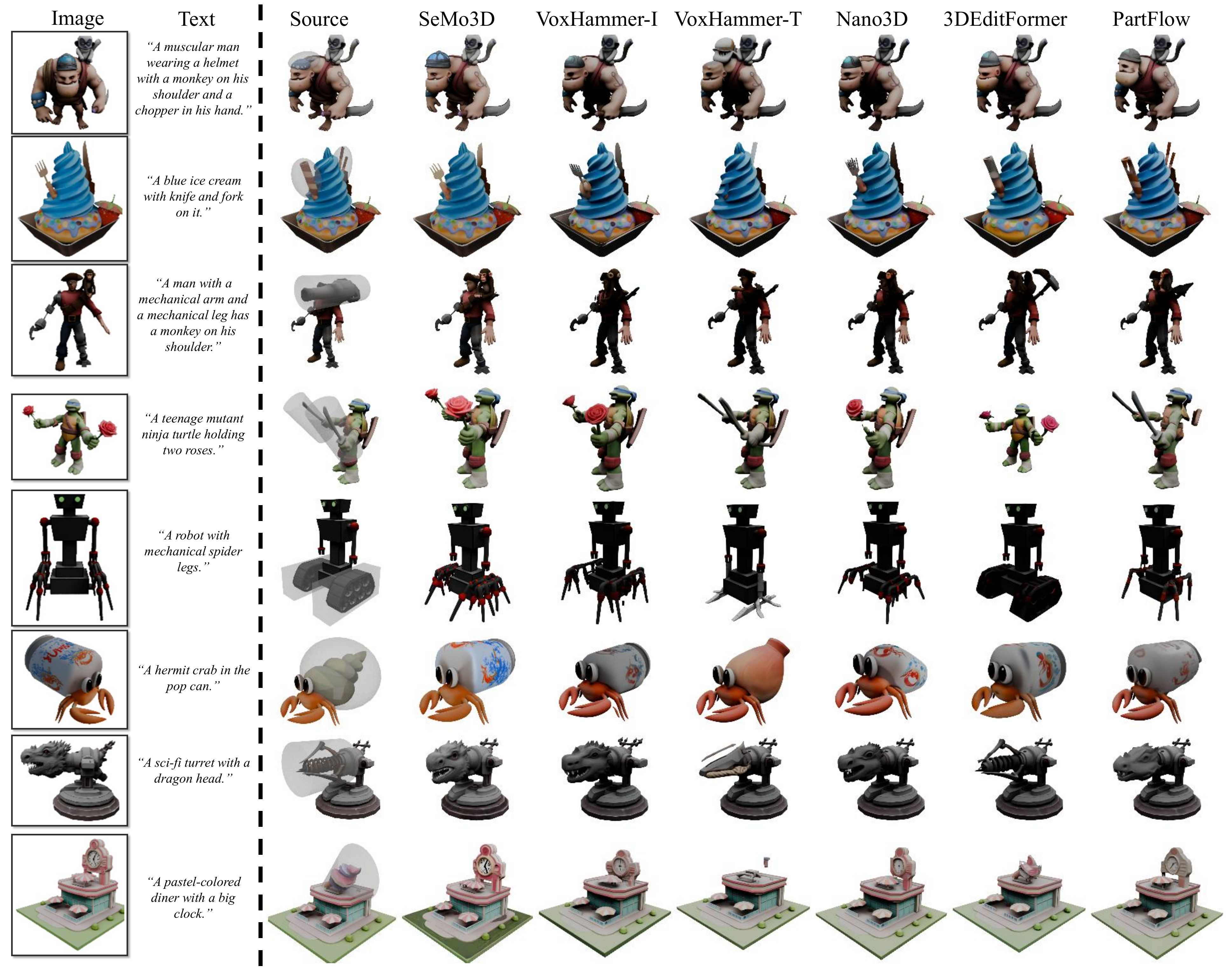}
    \caption{Additional qualitative comparison of 3D editing on Edit3D-Bench~\cite{voxhammer}. Translucent volumes mark the edit regions.}
    \label{fig:supp_edit}
\end{figure*}

\begin{figure*}[tp]
    \centering
    \includegraphics[width=0.95\linewidth]{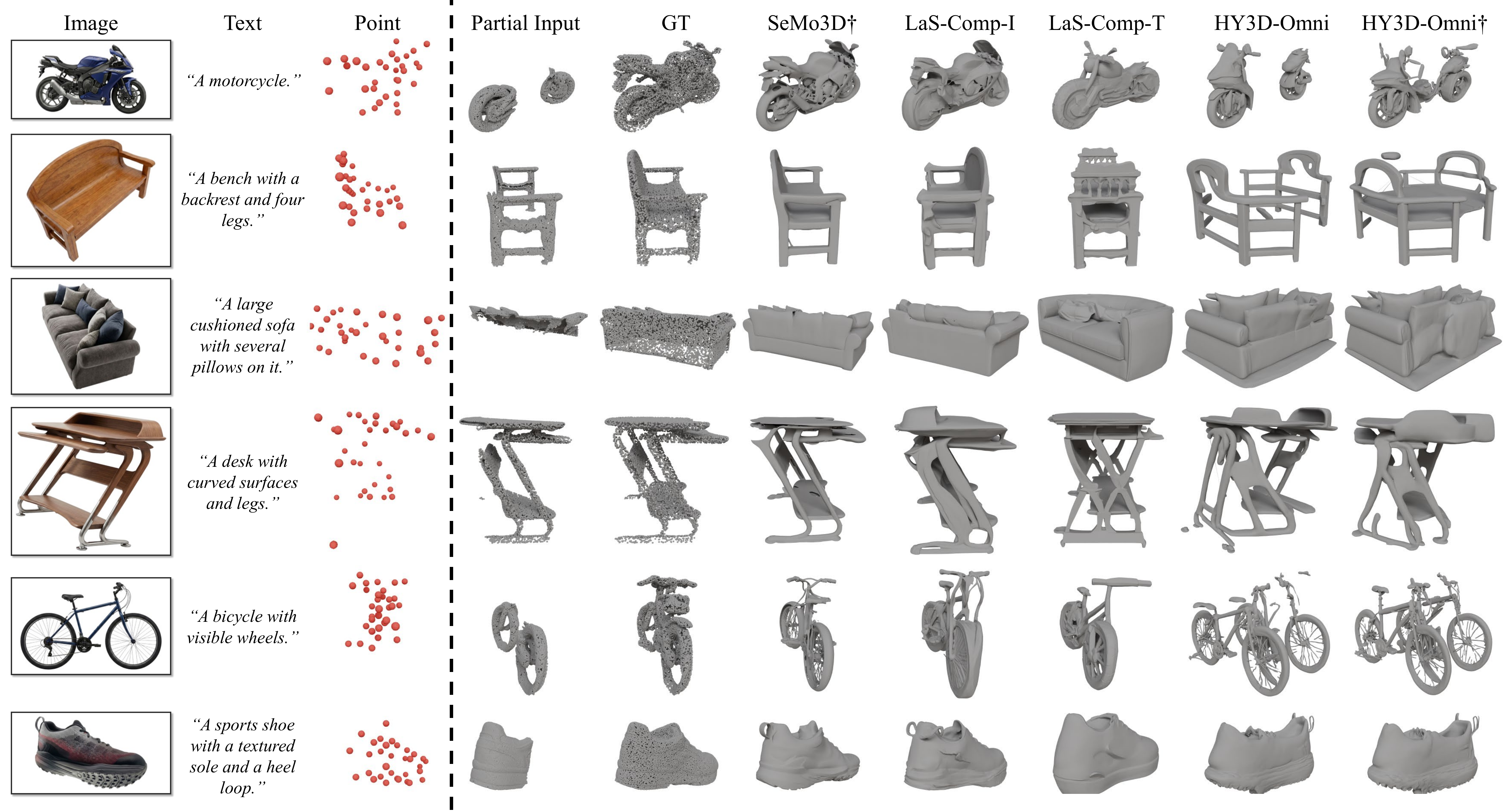}
    \caption{Additional qualitative comparison of 3D completion on Omni-Comp~\cite{lascomp}.}
    \label{fig:supp_comp}
\end{figure*}

\noindent\textbf{3D Generation, Editing, and Completion.}
\cref{fig:supp_gen,fig:supp_edit,fig:supp_comp} show additional comparisons.
In generation, SeMo3D recovers parts hidden from the input view, such as the stein handle, which most image-only baselines miss.
In editing, it modifies only the target region while preserving the rest, as with the rose-holding turtle.
In completion, it retains the partial input and fills the missing parts coherently, whereas HY3D-Omni may duplicate structures, such as the extra bicycle frame.

\begin{figure*}[tp]
    \centering
    \includegraphics[width=0.95\linewidth]{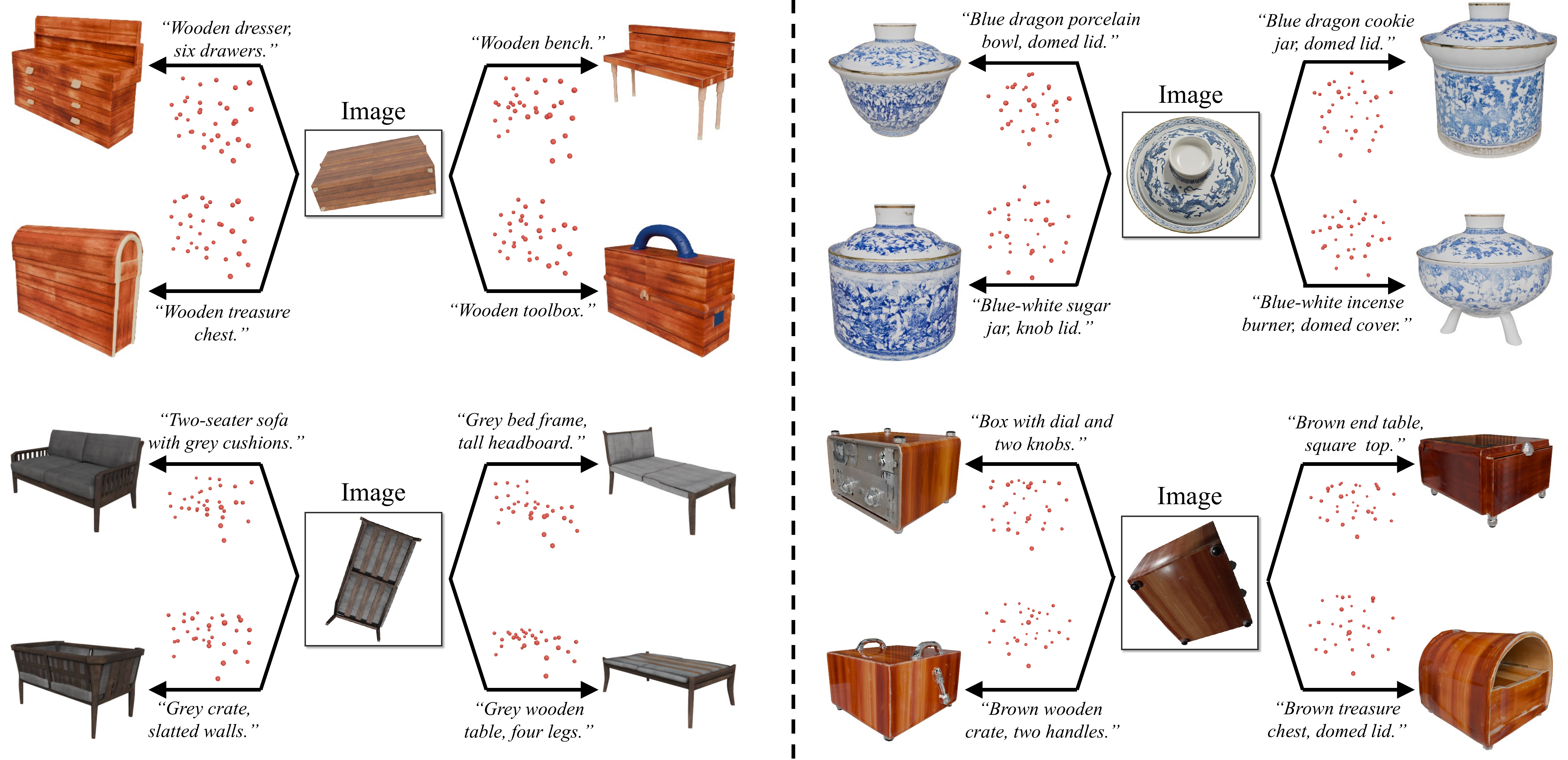}
    \caption{More multimodal control results.}
    \label{fig:supp_multicond}
\end{figure*}

\noindent\textbf{Multimodal Control.}
\cref{fig:supp_multicond} further shows that, with a fixed image, varying text and points produces diverse assets faithful to the image appearance.

\end{document}

%% file: preamble.tex
\usepackage{multirow}
\usepackage{makecell}
\usepackage{arydshln}
\usepackage{graphicx}
\usepackage{pifont}
\usepackage{cuted}
\usepackage{amsmath}
\usepackage{algorithm}
\usepackage{algpseudocode}
\usepackage{amsfonts}
\usepackage{amssymb}
\usepackage{dblfloatfix}
\usepackage[accsupp]{axessibility}  

